%% file: CVPR_camera_ready.tex
\documentclass[10pt,twocolumn,letterpaper]{article}

\usepackage[pagenumbers]{cvpr} 


\definecolor{cvprblue}{rgb}{0.21,0.49,0.74}
\usepackage[pagebackref,breaklinks,colorlinks,allcolors=cvprblue]{hyperref}

\input{include/packages}

\input{include/commands}

\def\paperID{3892} 
\def\confName{CVPR}
\def\confYear{2025}

\title{Hyperdimensional Uncertainty Quantification for Multimodal Uncertainty Fusion in Autonomous Vehicles Perception}

\author{Luke Chen\thanks{Corresponding Author. 
Distribution A: Approved for public release; distribution unlimited. OPSEC~\#~9537}
\hspace{0.25em} Junyao Wang 
\hspace{0.25em} Trier Mortlock 
\hspace{0.25em} Pramod Khargonekar 
\hspace{0.25em} Mohammad Abdullah Al Faruque \\
University of California, Irvine \\
{\tt\small \{panwangc, junyaow4, tmortloc, pramod.khargonekar, alfaruqu\}@uci.edu
} \\
}

\begin{document}
\maketitle
\input{chapters/0_abstract}
\input{chapters/1_intro}
\input{chapters/2_related}

\input{chapters/3_method}
\input{chapters/4_exp}
\input{chapters/5_conclusion}
{
    \small
    \bibliographystyle{ieeenat_fullname}
    \bibliography{reference}
}

\input{include/appendix}

\end{document}

%% file: include/packages.tex
\usepackage{graphicx}
\usepackage[utf8]{inputenc} 
\usepackage[T1]{fontenc}    
\usepackage{hyperref}       
\usepackage{url}            
\usepackage{booktabs}       
\usepackage{amsfonts}       
\usepackage{nicefrac}       
\usepackage{microtype}      
\usepackage{xcolor}         
\usepackage{xcolor,colortbl, soul}
\usepackage{xspace}
\usepackage{arydshln}

\usepackage{amsmath}
\usepackage{bbm}

\usepackage{multirow}

\usepackage[accsupp]{axessibility}  

%% file: include/commands.tex
\newcommand{\ourmethod}{\textit{{HyperDUM}}\xspace}

\newtheorem{corollary}{Corollary}

\newtheorem{definition}{Definition}

\newcommand{\reducedstrut}{\vrule width 0pt height .9\ht\strutbox depth .9\dp\strutbox\relax}

\newcommand{\bluehl}[1]{%
  \begingroup
  \setlength{\fboxsep}{0pt}%
  \colorbox{blue!25}{\reducedstrut#1\/}%
  \endgroup
}
\newcommand{\lightbluehl}[1]{%
  \begingroup
  \setlength{\fboxsep}{0pt}%
  \colorbox{blue!12}{\reducedstrut#1\/}%
  \endgroup
}

%% file: chapters/0_abstract.tex
\begin{abstract}
Uncertainty Quantification (UQ) is crucial for ensuring the reliability of machine learning models deployed in real-world autonomous systems. 
However, existing approaches typically quantify task-level output prediction uncertainty without considering epistemic uncertainty at the multimodal feature fusion level, leading to sub-optimal outcomes.
Additionally, popular uncertainty quantification methods, e.g., Bayesian approximations, remain challenging to deploy in practice due to high computational costs in training and inference. 
In this paper, we propose \ourmethod, a novel deterministic uncertainty method (DUM) that efficiently quantifies feature-level epistemic uncertainty by leveraging hyperdimensional computing.
Our method captures the channel and spatial uncertainties through channel and patch -wise projection and bundling techniques respectively.
Multimodal sensor features are then adaptively weighted to mitigate uncertainty propagation and improve feature fusion.
Our evaluations show that \ourmethod on average outperforms the state-of-the-art (SOTA) algorithms by up to 2.01\%/1.27\% in 3D Object Detection and up to 1.29\% improvement over baselines in semantic segmentation tasks under various types of uncertainties.
Notably, \ourmethod requires $2.36\times$ less Floating Point Operations and up to $38.30\times$ less parameters than SOTA methods, providing an efficient solution for real-world autonomous systems.
\end{abstract}

%% file: chapters/1_intro.tex
\section{Introduction}\label{sec:intro}
Effectively quantifying and addressing \textit{uncertainty} plays a critical role in ensuring the reliability of ML models in real-world applications, especially where the decisions are consequential, e.g., autonomous driving.
Models failing to address uncertainty can produce overconfident outputs, leading to compromised performance \cite{postels2019sampling,fervers2023uncertainty,hu2023planning,wang2024rs2g}.
In light of autonomous systems, uncertainties can arise from adverse weather conditions, sensor failures or noises, environment dynamics (movements of other actors, road conditions, vehicle dynamics), and various corner cases \cite{blum2019fishyscapes,goan2023uncertainty,mukhoti2023deep,mortlock2024castnet}.
Modern autonomous systems leverage multimodal sensor fusion to broaden the information scope and learn comprehensive representations of the surrounding environment~\cite{fayyad2020deep,yeong2021sensor,malawade2022hydrafusion}.
However, commonly employed sensors such as cameras, lidar, and radar, exhibit varying degrees of robustness and vulnerability across different driving scenarios~\cite{wang2022adversarial,vargas2021overview}.
For instance, cameras are crucial in delivering rich perceptual information while they are particularly vulnerable to lighting variations and occlusions~\cite{zhu2023understanding}.
Lidar and radar are more robust to visual perturbations but can be affected by multi-path interference and occlusion~\cite{bhupathiraju2023emi,li2020lidar}.
Therefore, it is important to consider the uncertainty of each modality before the fusion of multimodal features to achieve optimal results~\cite{heidecker2021application}.

Popular \textit{Uncertainty Quantification} (UQ) methods include deep ensembles, Bayesian neural networks, and Bayesian approximation methods such as Monte Carlo dropout \cite{lakshminarayanan2017simple, kwon2020uncertainty, gal2016dropout}.
However, these methods often introduce large training and/or inference overheads compounded by the growing complexity of modern Deep Neural Networks (DNN), ultimately challenging their practicality under real-time computing constraints~\cite{dusenberry2020efficient, jia2020efficient}.
Other lines of works such as Conformal Prediction (CP) and Posterior Networks (PostNets) based on evidential deep learning either cannot quantify feature-level epistemic uncertainty or require vast amounts of quality data to accurately estimate the posteriors \cite{angelopoulos2023conformal, charpentier2020posterior}.
Deterministic Uncertainty Methods (DUMs), aiming to efficiently quantify uncertainty using only a single forward pass \cite{postels2021practicality, mi2022training, franchi2022latent}, have become an emerging trend.
DUMs rely on statistical/morphological characteristics of latent features to quantify epistemic uncertainty.
For instance, \cite{franchi2022latent} employed Prototype Learning (PL) techniques to capture the ``knowledge'' of DNNs by constructing a ``memory bank'' of prototypes to predict uncertainty.

To address these problems, we propose \ourmethod, a novel deterministic uncertainty method for efficient quantification of feature-level epistemic uncertainty.
In contrast to existing methods, \ourmethod projects latent features into hyperdimensional space to form hyperdimensional prototypes, and then estimates uncertainty based on the distance between features of unseen samples and these prototypes.

To the best of our knowledge, \ourmethod is the first deterministic uncertainty quantification method utilizing hyperdimensional feature prototyping with \textbf{Channel-wise} and \textbf{Patch-wise} projection and bundling to deliver accurate and efficient epistemic uncertainty quantification of latent features.
Detailed experimental evaluations on the DeLiVER and aiMotive datasets for semantic segmentation and object detection tasks demonstrate that \ourmethod can efficiently and effectively quantify feature uncertainty and improve fusion performance when compared to standard Bayesian approximation inference baselines and SOTA methods.

%% file: chapters/2_related.tex
\section{Related Works}\label{sec:related}
\subsection{Uncertainty Quantification}
Predictive uncertainty is composed of three factors, model/epistemic uncertainty, data/aleatoric uncertainty, and distributional uncertainty \cite{malinin2018predictive}.
Epistemic uncertainty, stemming from the word epistemology which is the study of knowledge, is associated with a model’s lack of knowledge of underrepresented data. 
The scope of this work is to efficiently quantify epistemic uncertainty at the latent feature level to enable uncertainty-weighting prior to feature fusion.

Existing UQ approaches are typically derived from Bayesian principles, including Bayesian neural networks which represent model weights as probability distributions, deep ensembles which sample output distributions from diverse models \cite{lakshminarayanan2017simple}, and Monte Carlo dropout which is similar to deep ensembles but utilizes a single model with dropout layers \cite{gal2016dropout}.
However, these approaches remain challenging in practice due to enormous training and inference costs~\cite{dusenberry2020efficient, jia2020efficient}.
Conformal prediction (CP) is a model-agnostic UQ technique that provides prediction intervals with guaranteed coverage, ensuring the true value falls within the interval at a specified probability \cite{angelopoulos2023conformal}.
However, while effective in producing reliable task-level uncertainties, CP does not capture uncertainties at intermediate model abstractions, a limitation especially evident in multimodal fusion models, where uncertainty propagates across different modalities before the final output.
Evidential deep learning is another line of work, i.e. Posterior Networks which aim to quantify both aleatoric and epistemic uncertainty by directly estimating a posterior distribution over the model's outputs/predictions but is sensitive to data quality for accurately estimating the posteriors \cite{charpentier2020posterior}.
Deterministic uncertainty quantification is a growing research area that aims to efficiently compute epistemic uncertainty of DNNs with a single forward pass~\cite{postels2021practicality, mi2022training, franchi2022latent}.
For instance, \cite{mi2022training} is a training-free method that injects Gaussian noise into intermediate layers to compute the variance across the representations as a measure of uncertainty.
Latent Deterministic Uncertainty (LDU) constructs prototypes from latent features by enforcing dissimilarity between prototypes and entropy between latent features while correlating uncertainty and the downstream task~\cite{franchi2022latent, mukhoti2023deep,she2021dive}.
Recent research~\cite{ni2023brain} demonstrated that constructing hyperdimensional prototypes can provide comparable performance to existing aleatoric uncertainty quantification solutions, e.g., Bayesian approximation, and offer significant speedups in training and inference. 
However, it is applied to regression tasks for aleatoric uncertainty.
In contrast, our work quantifies epistemic uncertainty on the latent features, extending its applicability to multimodal fusion models and complex tasks such as object detection and semantic segmentation.

\subsection{Multimodal Uncertainty-Aware Fusion}
Multimodal sensor fusion aims to provide more comprehensive representations and address the limitations of unimodal sensor failure cases~\cite{yeong2021sensor,fayyad2020deep, malawade2022hydrafusion}.
Existing approaches dealing with uncertainties fall into two categories: feature-level fusion and output-level fusion. 
Feature-level fusion, also known as intermediate fusion, applies an uncertainty weighting that scales the features to minimize uncertainty~\cite{singh2023transformer, liang2018deep}. 
For instance, \cite{han2022multimodal} computes feature-informativeness and scales the original features based on informativeness.
On the other hand, output-level fusion, also known as late-fusion, uses uncertainty to calibrate the outputs of the model, i.e. calibrating softmax probabilities~\cite{pereira2023comparing}.
In contrast to existing uncertainty-based fusion works that commonly rely on Bayesian approximated uncertainty fusion, our proposed \ourmethod is the first to investigate deterministic uncertainty fusion methods in a multimodal autonomous systems setting.

%% file: chapters/3_method.tex
\section{Methodology}\label{sec:method}
\begin{figure*}[!ht]
\centering
\includegraphics[width=1\textwidth]{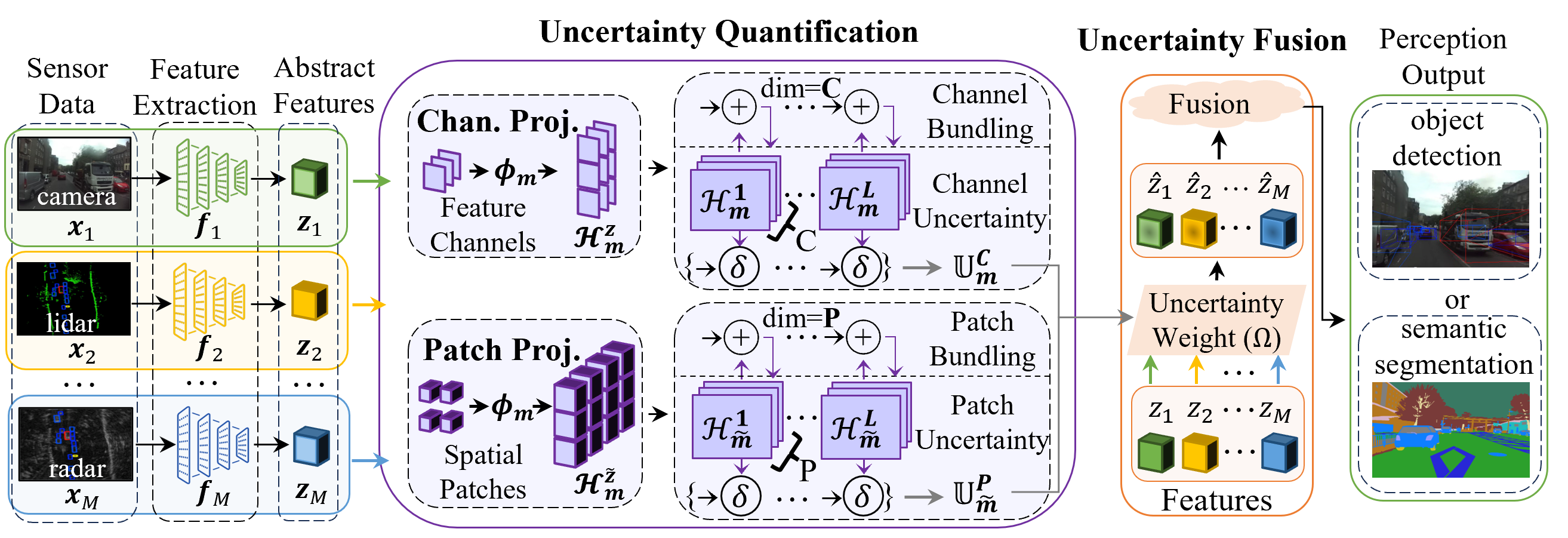}
\caption{Multimodal model with uncertainty quantification and uncertainty fusion for autonomous driving perception tasks.}
\label{fig:proposed}
\end{figure*}
\subsection{Preliminaries}
Given an input sample ${\bf{x}}= \{x_{1}, x_{2}, \cdots, x_{M}\}$ consisting of $M$ multimodal sensor inputs.
We have a multimodal model $\mathbb M$
with pre-trained feature encoders ${\bf{f}} = \{f_{1}, \cdots, f_{M}\}$ that extract modality-specific abstract features~\footnote{Cross-modal feature alignment is not required but may be enforced by the backbone.} $z_{m} = f_{m}(x_m)$ where $1\leq m \leq M$.
Additionally, the model contains fusion blocks 
$\mathcal{F}(z_{i},z_{j})$ where $i,j \in \{1, \cdots, M\}$ such that $i\neq j$ which combine modality-specific features to learn cross-modal fusion features $g$.
Assuming there are $K$ fusion blocks, the fused features ${\bf{g}} = \{g_{1}, \cdots, g_{K}\}$ are passed to the task head $h(\mathrm{g})$ to output the corresponding task predictions $y$ (i.e. bounding boxes and object classification for object detection, and pixel-wise class labels for semantic segmentation).
To account for feature uncertainty, an uncertainty quantification module $\mathbb{U}_{m}(z_{m})$ is inserted immediately following each feature extractor to predict an uncertainty value $u_{m}$.
Given the uncertainty value, a learnable uncertainty weighting module $\Omega(z_{m}, u_{m})$ reweights the input features to produce an uncertainty-aware feature $\hat{z}_{m}$ for each modality.
Thus, the input to the fusion blocks now becomes $\mathcal{F}(\hat{z_{i}},\hat{z_{j}})$ to produce uncertainty-weighted fusion features $\hat{\mathrm{g}}$ which are forwarded to be fused and subsequently used by the task head $h(\mathrm{\hat{g}})$ to produce the outputs.
The system flow of the described multimodal fusion model with uncertainty quantification is illustrated in Figure~\ref{fig:proposed}.

\subsubsection{Vector Symbolic Architectures/Hyperdimensional Computing}
Vector Symbolic Architectures (VSA) / Hyperdimensional Computing (HDC) are a class of algorithms for creating and handling hyperdimensional representations to mimic cognitive functions \cite{levy2008vector, schlegel2022comparison}.
VSA starts with mapping input data to hyperdimensional vectors, or hypervectors $\mathcal{H}$, through a projection function $\phi: \mathcal{Z} \rightarrow \mathcal{\textbf{H}}$, where $\mathcal{Z} \in \mathbb{R}^n$ and $\mathcal{\textbf{H}} \in \mathbb{R}^d$ ($d\gg n$) is the hypervector space.
The projection function depends on the input data representation. 
This work assumes Euclidean input space.
For non-euclidean see~\cite{nunes2022graphhd}.
Following the projection, common operations on hypervectors include similarity calculation, {bundling}, and {binding}. 
\textbf{Similarity} ($\delta(\cdot, \cdot)$) calculation measures the distance between two hypervectors. For real-valued hypervectors, a common measure is cosine similarity. 
\textbf{Bundling} ($\bigoplus$) is an element-wise addition of hypervectors, e.g., ${\mathcal{H}}^{bundle}= {\mathcal{H}}^1 \bigoplus {\mathcal{H}}^2$, generating a hypervector with the same dimension as inputs.
\textbf{Binding} ($\bigotimes$) is an element-wise multiplication associating two hypervectors to create another near-orthogonal hypervector, e.g., ${\mathcal{H}}^{bind} ={\mathcal{H}}^{1}\bigotimes {\mathcal{H}}^{2}$ 
 where $\delta({\mathcal{H}}^{bind}, {\mathcal{H}}^{1})\approx 0$ and $\delta({\mathcal{H}}^{bind}, {\mathcal{H}}^{2})\approx 0$. 

Our work mainly leverages the \textbf{bundling} operation of VSA for the formation of prototypes.
In high-dimensional space, bundling mimics how human brains \textit{memorize} information \cite{yu2022understanding,thomas2021theoretical}.
For instance, for ${\mathcal{H}}^{bundle}= {\mathcal{H}}^1 +{\mathcal{H}}^2$, we have $\delta({\mathcal{H}}^{bundle},{\mathcal{H}}^1)\gg 0 $ while $\delta({\mathcal{H}}^{bundle},{\mathcal{H}}^3)\approx 0 $ (${\mathcal{H}}^3\neq {\mathcal{H}}^1, {\mathcal{H}}^2)$. 
{In other words, VSA can generate prototypical representations by bundling the high-dimensional vectors of individual data samples, making it especially useful for prototype learning.}
Specifically, suppose we have labeled samples $\mathcal D = \{(z^{1}_{1},y^{1}_{1}), (z^{2}_{1},y^{2}_{1}), \ldots, (z^{N}_{M},y^{N}_{M})\}$ where $z^{i}_{m} \in \mathcal{Z}$ and $y^{i}_{m} \in \{l\}^{L}_{l=1}$, we can form $L$ prototypes by bundling hypervectors corresponding to a particular label:
\begin{equation}\label{eqn:hdp}
    \mathcal{H}^{l}_m = \bigoplus_{i:y^{i}={l}}\phi_{m}(z^{i}_{m})
\end{equation}
where each hyperdimensional prototype can represent an aspect of the data samples $\mathcal D$ depending on the label definition, i.e. class label (cat and dog) or context label (weather conditions which we use in this work).
We then define the notion of uncertainty as the set of similarity between hypervectors and hyperdimensional prototype of modality $m$ as follows:
\begin{align}\label{eqn:uncert1}
\mathbb{U}_{m} = \bigcup_{l=1}^L\{\delta(\mathcal{H}^{z}_{m}, \mathcal{H}^{l}_{m})\}, \text{where}~|\mathbb{U}_{m}|=L
\end{align}

where $\mathbb{U}_{m}$ is the set of similarity distances between a hypervector $\mathcal{H}^{z}_{m}$ and each hyperdimensional prototype $\mathcal{H}^{l}_{m}$, $l= 1, 2, \ldots, L$ and used to express the similarity uncertainty \footnote{We derive uncertainty quantifiability from similarity/expressivity of VSAs/HDCs based on recent theorems in the Appendix.}.
The notion of similarity for uncertainty has been used in other prototype learning methods \cite{li2024prototype, franchi2022latent}, where the former assigns the variations of similarities between an instance and the prototypes as the belief masses and the latter utilizes similarity to optimize the uncertainty loss.
\subsection{\ourmethod}
Following the conventions of VSA for projection and bundling, we propose two modifications for uncertainty quantification: \textbf{1) Channel-wise and 2) Patch-wise Projection \& Bundling (CPB/PPB).}
CPB modifies the projection and bundling operations to capture the uncertainty of each channel per modality separately, whereas PPB allows the VSA to quantify the spatial uncertainty without losing fine-grained spatial information through patching.
Figure~\ref{fig:proposed} visualizes the CPB and PPB operations of the proposed method.

\subsubsection{Channel-wise Projection \& Bundling (CPB)}
Latent features are generally multi-channeled.
By convention prototype formation based on Equation \ref{eqn:hdp} projects and bundles all feature dimensions onto a common representation space $\mathcal{\textbf{H}}$ \cite{wilson2023hyperdimensional, wang2023disthd}. 
Given latent feature $z^{i}_{m}\in\mathbb{R}^{C\times H\times W}$ with $C$ denoting feature channels and $H, W$ the spatial dimensions, adaptive pooling (max or average) is first used to reduce the spatial dimensions to $z^{pooled}_{m} \in \mathbb{R}^{C}$.
Through matrix multiplication and proper dimension initialization of the projection matrix $\Phi$, the correct hypervector dimensions can be obtained for $\mathcal{H}^{z}_{m}$ as follows:
\begin{align}\label{eqn:matmul}
    \phi_{m}(z^{i}_{m}) = \Phi^{(d\times C)} \cdot z^{pooled(C)}_{m} = {\mathcal{H}^{z}_{m}}^{(d)}
\end{align}
Another method to handle multi-channels inputs is by projecting each feature dimension separately followed by binding and subsequent bundling to a common space \cite{thomas2021theoretical}.
However, we note that the above operations neglect the fact that different feature channels learn different aspects of the input and may contribute more or less to the overall uncertainty.
For example, \cite{li2022uncertainty} showed that sharing the same uncertain distribution among different channels is less effective for out-of-distribution generalization, while considering each channel uncertainty separately brings better performances due to the different channel potentials.
Therefore, instead of absorbing the channel dimensions during projection and bundling, we retain them as follows:
\begin{align}\label{eqn:matmul_ch}
    \phi_{m}(z^{i}_{m}) = \Phi^{(d\times C)} \otimes z^{pooled(C)}_{m} = {\mathcal{H}^{z}_{m}}^{(d\times C)}
\end{align}
where we use the Einstein summation notation $\otimes$ (not the binding $\bigotimes$ notation) to retain the channel dimension.
Consequently, the bundling operation is modified to account for the channel dimensions:
\begin{equation}\label{eqn:chdp}
    \mathcal{H}^{l}_m = \bigoplus^{dim=C}_{i:y^{i}={l}}\phi_{m}(z^{i}_{m})
\end{equation}
where $dim=C$ indicates the dimension over which the element-wise addition is performed and $\mathcal{H}^{l}_m$ is now $\mathbb{R}^{d\times C}$.
Additionally, the similarity uncertainty now outputs similarities for each channel dimension by representing Equation~\ref{eqn:uncert1}:
\begin{align}\label{eqn:uncert1_chan}
\mathbb{U}^{C}_{m} = \bigcup_{c=1}^C\{\bigcup_{l=1}^L\{\delta(\mathcal{H}^{z}_{m}[c], \mathcal{H}^{l}_{m}[c])\}\}, \text{s.t.}~|\mathbb{U}^{C}_{m}|=C\times L
\end{align}

We empirically demonstrate through experiments that channel projection and bundling outperforms conventional projection and bundling for the task of multimodal uncertainty-aware feature fusion.

\subsubsection{Patch-wise Projection \& Bundling (PPB)}\label{sec:ppb}
Channel projection and bundling enables our method to capture the holistic uncertainties of each feature channel.
However, to capture the finer-granularity uncertainties of the spatial dimensions requires a different approach.
By convention spatial dimensions are either pooled to a single value and projected to high dimensions as shown by Equation~\ref{eqn:matmul} or each spatial dimension is separately projected to high dimensions.
The former loses spatial information due to pooling and the latter is computationally infeasible and memory intensive given the large spatial dimensions of deep neural networks.

Inspired by the idea of using image patches to capture spatial information in anomaly detection and image classification~\cite{dosovitskiy2020image, roth2022towards}, we propose PPB which projects and bundles spatial patches of latent features to hyperdimensional space to capture the spatial uncertainties.
Specifically, given latent feature $z^{i}_{m}\in\mathbb{R}^{C\times H\times W}$, we slice $(H,W)$ into $P$ patches of resolution $(\tilde{h}$,$\tilde{w})$ where $\tilde{h} = H/\sqrt(P)$, $\tilde{w} = W/\sqrt(P)$.
This gives us the set of patched features $\tilde{z}^{i}_{m} = \{\tilde{z}^{i,j}_{m}\}^P_{j=1}$ where each $\tilde{z}^{i,j}_{m}\in\mathbb{R}^{C\times \tilde{h}\times \tilde{w}}$ are pooled to $\mathbb{R}^{C}$ and form the set of pooled patched features $\tilde{z}^{pooled}_{m}\in\mathbb{R}^{C\times P}$.

Given patched features $\tilde{z}^{pooled}_{m}$, we project them according to Equation~\ref{eqn:matmul} to form the set of patched projections:

\begin{align}\label{eqn:matmul_patched}
    \phi_{m}(\tilde{z}^{i}_{m}) = \Phi^{(d\times C)} \cdot \tilde{z}^{pooled(C\times P)}_{m} = {\mathcal{H}^{\tilde{z}}_{m}}^{(d\times P)}
\end{align}

Consequently, these projections are bundled for each patch according to Equation~\ref{eqn:hdp} to produce the set of patched prototypes as follows:

\begin{equation}\label{eqn:hdp_patched}
    \mathcal{H}^{l}_{\tilde{m}} = \bigoplus^{dim=P}_{i:y^{i}={l}}\phi_{m}(\tilde{z}^{i}_{m})
\end{equation}

where $\mathcal{H}^{l}_{\tilde{m}}$ represents the patched prototypes of modality $m$, for each patch and for each context label $i$, making $|\mathcal{H}^{l}_{\tilde{m}}| = d\times P$.

Finally, we can obtain the spatial uncertainties for each patch of the input projection $\mathcal{H}^{\tilde{z}}_{m}$ by modifying Equation~\ref{eqn:uncert1}:

\begin{align}\label{eqn:uncert1_spat}
\mathbb{U}^{P}_{\tilde{m}} = \bigcup_{p=1}^P\{\bigcup_{l=1}^L\{\delta(\mathcal{H}^{\tilde{z}}_{m}[p], \mathcal{H}^{l}_{\tilde{m}}[p])\}\}, \text{s.t.}~|\mathbb{U}^{P}_{\tilde{m}}|=P\times L
\end{align}

Our final proposed approach combines both channel and patch methods for projection and bundling to capture channel-wise feature uncertainties, as well as finer-granularity spatial feature uncertainties, to provide a holistic epistemic uncertainty quantification of each modality.

%% file: chapters/4_exp.tex
\begin{figure*}[!ht]
\centering
\includegraphics[width=1\textwidth]{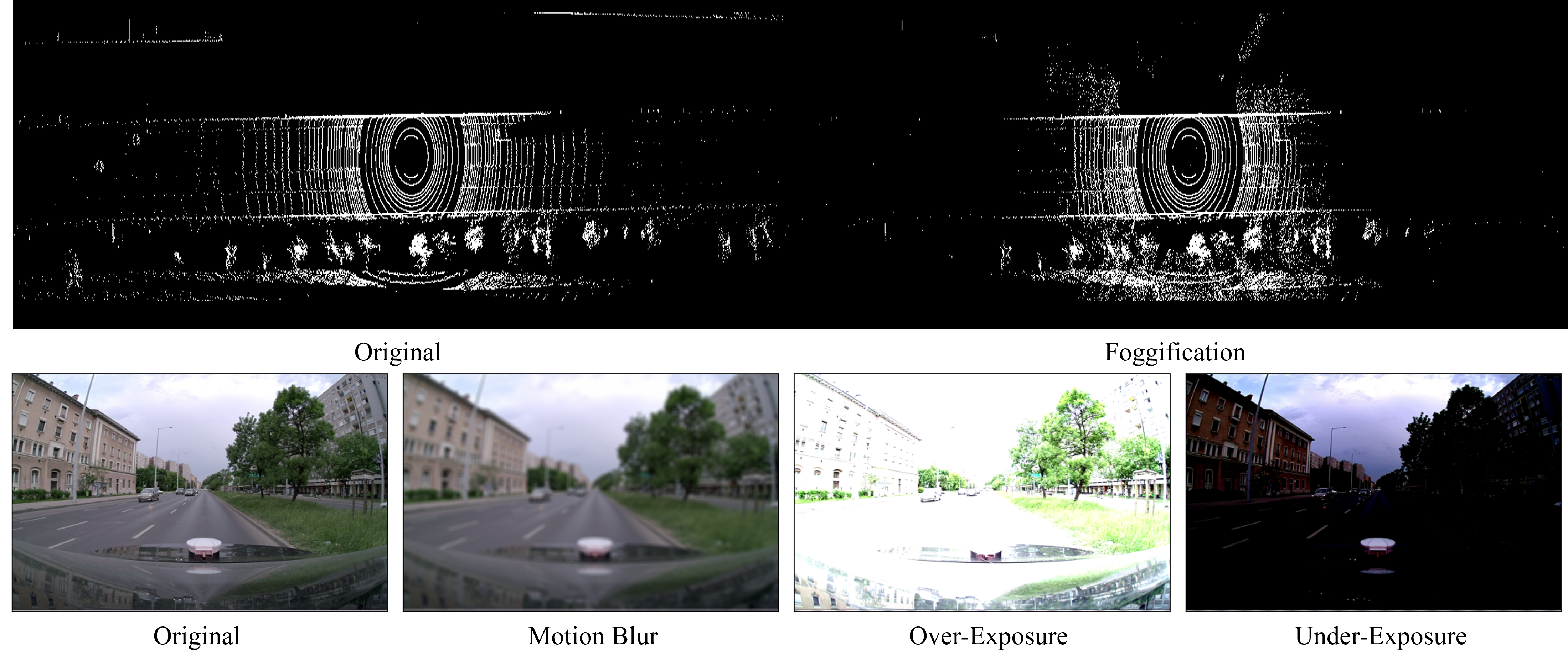}
\caption{Corner cases for the aiMotive 3D Object Detection dataset. (Top-Left) Normal Lidar point-cloud birds-eye-view. (Top-Right) Lidar Foggification (LF). (Bottom) In order from left to right (Normal, Motion Blur (MB), Over-Exposure (OE), Under-Exposure (UE))}
\label{fig:injections}
\end{figure*}
\section{Experiments}\label{sec:exp}
\subsection{Experimental Setup}
\subsubsection{Datasets}
We evaluate the effectiveness of \ourmethod for uncertainty-aware feature fusion using two multimodal autonomous driving datasets.
The aiMotive dataset \cite{matuszka2022aimotive} includes camera, lidar, and radar sensors for long-distance ($\geq 75m$) 3D object detection and features diverse scenes across various locations, times, and weather conditions.
The DeLiVER dataset \cite{zhang2023delivering} contains camera, lidar, event, and depth sensors for semantic segmentation, including adverse corner cases such as motion blur and lidar jitter.
Additionally, we integrate corner cases into the aiMotive dataset by applying various lighting effects to camera sensors, causing reduced clarity and scene obstruction.
These effects are created through image transformations such as GaussianBlur and exposure adjustments \cite{van2014scikit, paszke2017automatic}.
We also implement foggification \cite{bijelic2020seeing} to simulate synthetic fogging of lidar data, resulting in noisy point clouds with reduced field of view.
These corner case effects can be visualized in Figure~\ref{fig:injections}. Details in the appendix.
\subsubsection{Models}
We utilize the pretrained BEVFusion \cite{matuszka2022aimotive} and CMNeXt \cite{zhang2023delivering} models as the fusion architectures for aiMotive and DeLiVER, respectively.
BEVFusion is built upon VoxelNet \cite{zhou2018voxelnet} and BEVDepth \cite{li2023bevdepth} to fuse camera and lidar representations on a unified BEV space.
CMNeXt is a four-stage pyramidal fusion model that treats the RGB representation as the primary branch and other modalities as the secondary branch.
We integrate the uncertainty quantification module (UQM)
at each model's respective pre-fusion points.
For aiMotive this is chosen to be at the feature concatenation step, right before the bev\_fuse step.
For DeLiVER, we insert the UQM immediately after the feature rectification module and right before the feature fusion module.
Given that the CMNeXT model has four stages, we instantiate one UQM per stage.
Details in the Appendix.
\subsubsection{Baseline Methods}
We implemented different UQ methods and compared their performances against our method.
For each implementation, we followed their open-sourced code implementation and hyperparameter settings.
Specifically, for Infer-dropout (InfMCD) and infer-noise (InfNoise), we followed their suggestions and chose 10 forward passes and used the variance of the outputs for uncertainty estimation.
In addition to variance, we computed predictive entropy, mutual information, and entropy as additional uncertainty estimates following \cite{tian2020uno}.
It was shown that using more uncertainty metrics can provide more effective quantification for different types of uncertainties.
For Posterior Network (PostNet), we constructed PostNets for each modality so that they can estimate the modality-specific uncertainty independently and be used in the uncertainty-aware feature fusion process. 
The use of multiple PostNets is similar to \cite{itkina2023interpretable} in their construction of separate PostNets for agent's past behavior, road structure map, and social context respectively.
For latent deterministic uncertainty (LDU), we set the number of learnable prototypes to the number of scenarios.
Specifically, for aiMotive, the scenarios are Highway, Urban, Night, and Rain, while for DeLiVER, the scenarios are Cloudy, Foggy, Night, Rainy, and Sunny. 
Finally, for all methods, all layers prior to the UQM are kept frozen and we fine-tune the uncertainty weighting layer, which takes the respective uncertainty metrics as input, along with other post-fusion layers, to enable the model to learn the uncertainty-weighted features.
Details of the fine-tuning procedures and visualizations of the architectures and UQM insertion points are provided in the Appendix.
In all of our experiments the best one is highlighted and bolded and the 2nd best is lightly-highlighted.

\subsection{Multimodal Uncertainty Fusion for Autonomous Vehicles Perception}

\subsubsection{3D Object Detection}
\begin{table}[!ht]
    \scriptsize
    \centering
    \addtolength{\tabcolsep}{-0.6em}{
    \begin{tabular}{l|c|c|c|c|c}
        \hline
        \textbf{UQ Method} & \textbf{Highway} & \textbf{Urban} & \textbf{Night} & \textbf{Rain} & \textbf{Mean} \\
        \hline
        BEVFusion~\cite{matuszka2022aimotive}& \lightbluehl{72.55}/69.08 & \lightbluehl{63.72/63.52} & 74.74/73.13 & \lightbluehl{42.75/42.83} & \lightbluehl{65.67/64.79} \\
        InfMCD~\cite{mi2022training} & \bluehl{\textbf{72.58}/\textbf{69.65}} & 62.02/59.59 & 75.05/73.72 & 35.58/37.41 & 64.56/64.58 \\
        InfNoise \cite{mi2022training} & 70.11/68.52 & 63.10/60.42 & 75.08/74.19 & 35.58/37.41 & 64.59/64.56 \\
        PostNet~\cite{charpentier2020posterior} & 69.19/68.02 & 61.38/59.84 & 73.32/68.97 & 34.95/37.60 & 63.13/60.21 \\
        LDU~\cite{franchi2022latent} & 71.48/69.33 & 62.54/60.11 & \lightbluehl{76.49/75.04} & 40.49/41.24 & 64.69/64.73 \\
        \ourmethod & 72.23/\lightbluehl{69.58} & \bluehl{\textbf{64.77}/\textbf{64.48}} & \bluehl{\textbf{76.69}/\textbf{75.15}} & \bluehl{\textbf{44.78/45.48}} & \bluehl{\textbf{66.70}/\textbf{66.00}} \\
        \hline
    \end{tabular}
    }
    \caption{Results on aiMotive under diverse scenes.}
    \label{tab:aimotive}
\end{table}

Table~\ref{tab:aimotive} compares the all-point/11-point interpolation Average Precision (AP) metric for object detection under diverse scenes.
It can be seen that apart from the Highway scenario, \ourmethod outperforms all methods across all scenarios.
Specifically, \ourmethod improves upon the all-point AP by 2.01\% on average against the state-of-the-art (LDU).

\begin{table}[!ht]
    \scriptsize
    \centering
    \addtolength{\tabcolsep}{-0.6em}{
    \begin{tabular}{l|c|c|c|c|c}
        \hline
        \textbf{UQ Method} & \textbf{MB} & \textbf{OE} & \textbf{UE} & \textbf{LF} & \textbf{Mean} \\
        \hline
        BEVFusion~\cite{matuszka2022aimotive} & \lightbluehl{64.10/63.69} & 62.40/59.26 & \lightbluehl{63.57/63.25} & 65.07/65.08 & 63.79/62.75 \\
        InfMCD~\cite{mi2022training} & 62.00/61.99 & 64.47/64.28 & 63.25/63.49 & 64.27/63.81 & 63.50/\lightbluehl{63.39} \\
        InfNoise \cite{mi2022training} & 62.61/62.61 & 64.94/64.72 & 63.18/60.07 & 65.19/64.99 & \lightbluehl{63.98}/63.10 \\
        PostNet~\cite{charpentier2020posterior} & 60.42/58.90 & 63.12/59.99 & 60.20/59.25 & 64.13/60.60 & 61.97/59.69 \\
        LDU~\cite{franchi2022latent} & 62.06/62.22 & \lightbluehl{65.07/64.85} & 62.87/60.02 & \lightbluehl{65.47/65.10} & 63.87/63.05 \\
        \ourmethod & \bluehl{\textbf{64.39}/\textbf{63.99}} & \bluehl{\textbf{66.16}/\textbf{65.39}} & \bluehl{\textbf{64.18}/\textbf{63.91}} & \bluehl{\textbf{65.89}/\textbf{65.22}} & \bluehl{\textbf{65.16/64.62}} \\
        \hline
    \end{tabular}
    }
    \caption{Results on aiMotive under corner cases: Motion Blur (\textbf{MB}), Over-Exposure (\textbf{OE}), Under-Exposure (\textbf{UE}), LiDAR-Fog (\textbf{LF}). (Metrics: all-point AP/11-point interpolation AP)}
    \label{tab:aimotive_perb}
\end{table}

Table~\ref{tab:aimotive_perb} demonstrates the robustness of UQ methods under our injected corner cases.
These injections were unseen during training which complicates testing due to distribution shifts.
Each corner case is applied to and evaluated on the entire test set across all scenarios.
Whereas other methods fail and suffer performance degradations, \ourmethod only degrades 0.51\%/0.17\% compared to the baseline w/o injections demonstrating that \ourmethod is more robust even when facing distribution shifts.

\begin{table}[!ht]
    \scriptsize
    \centering
    \addtolength{\tabcolsep}{-0.6em}{
    \begin{tabular}{l|c|c|c|c|c}
        \hline
        \textbf{UQ Method} & \textbf{Highway} & \textbf{Urban} & \textbf{Night} & \textbf{Rain} & \textbf{Mean} \\
        \hline
        BEVFusion~\cite{matuszka2022aimotive} & \bluehl{\textbf{45.59}}/45.34 & \lightbluehl{46.38/45.53} & \bluehl{\textbf{44.49}}/\lightbluehl{45.87} & \bluehl{\textbf{42.65/42.93}} & \lightbluehl{45.84/46.87} \\
        InfMCD \cite{mi2022training} & \lightbluehl{45.57}/\bluehl{\textbf{46.29}} & 38.90/38.11 & 42.07/41.97 & 26.13/28.49 & 40.74/41.64 \\
        InfNoise~\cite{mi2022training} & 40.73/40.38 & 42.41/40.90 & 38.80/40.00 & 29.28/32.81 & 40.91/42.04 \\
        PostNet~\cite{charpentier2020posterior} & 41.13/40.64 & 39.47/40.16 & 40.37/41.92 & 28.56/28.96 & 39.37/41.66 \\
        LDU~\cite{franchi2022latent} & 43.07/45.01 & 40.17/41.71 & 42.90/42.52 & 31.12/34.08 & 40.65/42.17 \\
        \ourmethod & 44.86/\lightbluehl{46.06} & \bluehl{\textbf{47.89/48.91}} & \lightbluehl{44.42}/\bluehl{\textbf{46.12}} & \lightbluehl{40.54/41.63} & \bluehl{\textbf{46.52/48.27}}  \\
        \hline
    \end{tabular}
    }
    \caption{Results on aiMotive under diverse scenes in the distant region (>75m). (Metrics: all-point AP/11-point interpolation AP)}
    \label{tab:aimotive_lr}
\end{table}

Table~\ref{tab:aimotive_lr} is an experiment on the distant region setting, keeping only predictions and ground truths over $\geq 75m$.
As expected, all UQ methods experience major performance drops, especially for Rain.
This is likely due to the combined uncertainties of distance with rain obstruction causing substantial minimization of the uncertainty-weighted features (the weighting output is between 0-1 times the original feature value).
The original BEVFusion model maintains better performance because it is still able to ``guess'' the presence of objects as the features are not minimized by uncertainty.
									
\subsubsection{Semantic Segmentation}
\begin{table}[!ht]
    \scriptsize
    \centering
    \addtolength{\tabcolsep}{-0.5em}{
        \begin{tabular}{l|cccccc|c}
            \hline
            \multirow{3}{*}{\textbf{Scenarios}} & \multicolumn{7}{c}{\textbf{UQ Methods (Metrics: mIoU~$\uparrow$)}} \\
            \cline{2-8}
            & CMNeXt & InfMCD & 
            InfNoise & PostNet & LDU & Gemini & \ourmethod \\
            & \cite{zhang2023delivering} & \cite{mi2022training} & \cite{mi2022training} & \cite{charpentier2020posterior} & \cite{franchi2022latent} & \cite{jia2024geminifusion} & (Ours) \\
            \hline
            \textbf{Cloudy} & 68.70 & 69.23 & 69.21 & \lightbluehl{69.28} & 68.94 & -- & \bluehl{\textbf{69.76}}$^{+1.06}$\\
            \hline
            \textbf{Foggy} & 65.66 & \lightbluehl{66.18} & 66.10 & 66.04 & 65.72 & -- & \bluehl{\textbf{66.85}}$^{+1.19}$ \\
            \hline
            \textbf{Night} & 62.46 & \lightbluehl{63.44} & 63.14 & 62.97 & 63.06 & -- & \bluehl{\textbf{64.21}}$^{+1.75}$ \\
            \hline
            \textbf{Rainy} & 67.50 & 68.08 & 68.09 & \lightbluehl{68.16} & 67.82 & -- & \bluehl{\textbf{68.71}}$^{+1.21}$ \\
            \hline
            \textbf{Sunny} & 66.57 & 67.01 & \lightbluehl{67.16} & 66.93 & 66.75 & -- & \bluehl{\textbf{67.87}}$^{+1.30}$ \\
            \hline
            \hline
            \textbf{MB} & 62.91 & \lightbluehl{63.61} & 63.55 & 63.55 & 63.16 & -- & \bluehl{\textbf{64.28}}$^{+1.37}$ \\
            \hline
            \textbf{OE} & 64.59 & \lightbluehl{65.39} & 65.17 & 65.06 & 64.73 & -- & \bluehl{\textbf{65.67}}$^{+1.08}$ \\
            \hline
            \textbf{UE} & 60.00 & 60.38 & \lightbluehl{60.42} & 60.27 & 60.29 & -- & \bluehl{\textbf{61.20}}$^{+1.20}$ \\
            \hline
            \textbf{LJ} & 65.92 & 66.12 & 66.25 & 66.33 & \lightbluehl{66.40} & -- & \bluehl{\textbf{66.93}}$^{+1.01}$ \\
            \hline
            \textbf{EL} & 65.48 & 66.05 & \lightbluehl{66.17} & 66.06 & 65.89 & -- & \bluehl{\textbf{66.80}}$^{+1.32}$ \\
            \hline
            \hline
            \textbf{Mean} & 66.30 & \lightbluehl{66.90} & 66.86 & 66.77 & 66.60 & \lightbluehl{66.90} & \bluehl{\textbf{67.59}}$^{+1.29}$ \\
            \hline
        \end{tabular}
    }
    \caption{DeLiVER mean Intersection over Union (mIoU) performance under adverse weather and corner cases. Lidar-Jitter (\textbf{LJ}), Event Low-resolution (\textbf{EL}).}
    \label{tab:deliver_val_miou}
\end{table}
Table~\ref{tab:deliver_val_miou} shows the DeLiVER validation set performance on different scenarios and corner cases for the semantic segmentation task.
Overall, \ourmethod achieves the best performance over all methods across every weather and corner case scenario.
On average \ourmethod improves over the baseline by 1.29.
In particular, for the three most difficult scenarios including Under-Exposure, Night and Motion Blur, \ourmethod achieves at least 1.00 improvement over the baseline. 
For GeminiFusion~\cite{jia2024geminifusion}, we only report the mean as they omitted the detailed performance breakdown in their paper.
These results demonstrate that our method can better quantify the multimodal feature uncertainties and ultimately improve the fusion features for the semantic segmentation task.
Additionally, we note that all UQ methods improved over the baseline method.
This shows that incorporating UQ with fusion into multimodal pipelines not only enables UQ but also can improve baseline performance.

\begin{table}[!ht]
    \scriptsize
    \centering
    \addtolength{\tabcolsep}{-0.5em}{
        \begin{tabular}{l|ccccc|c}
            \hline
            \multirow{3}{*}{\textbf{Scenarios}} & \multicolumn{6}{c}{\textbf{UQ Methods (Metrics: ECE~$\downarrow$)}} \\
            \cline{2-7}
            & CMNeXt & InfMCD & 
            InfNoise & PostNet & LDU & \ourmethod \\
            & \cite{zhang2023delivering} & \cite{mi2022training} & \cite{mi2022training} & \cite{charpentier2020posterior} & \cite{franchi2022latent} & (Ours) \\
            \hline
            \textbf{Cloudy} & 1.18E-02 & 1.05E-02 & 1.07E-02 & \lightbluehl{0.99E-02} & 1.09E-02 & \bluehl{\textbf{0.94E-02}}$^{-0.24}$ \\
            \hline
            \textbf{Foggy} & 1.67E-02 & 1.48E-02 & 1.52E-02 & \lightbluehl{1.45E-02} & 1.56E-02 & \bluehl{\textbf{1.35E-02}}$^{-0.30}$ \\
            \hline
            \textbf{Night} & 1.88E-02 & \lightbluehl{1.63E-02} & 1.68E-02 & 1.71E-02 & 1.68E-02 & \bluehl{\textbf{1.39E-02}}$^{-0.49}$ \\
            \hline
            \textbf{Rainy} & 1.53E-02 & 1.37E-02 & 1.39E-02 & \lightbluehl{1.33E-02} & 1.43E-02 & \bluehl{\textbf{1.21E-02}}$^{-0.32}$ \\
            \hline
            \textbf{Sunny} & 1.42E-02 & 1.25E-02 & 1.27E-02 & \lightbluehl{1.24E-02} & 1.33E-02 & \bluehl{\textbf{1.12E-02}}$^{-0.30}$ \\
            \hline
            \hline
            \textbf{MB} & 1.54E-02 & \lightbluehl{1.32E-02} & 1.40E-02 & \lightbluehl{1.32E-02} & 1.44E-02 & \bluehl{\textbf{1.19E-02}}$^{-0.35}$ \\
            \hline
            \textbf{OE} & 1.49E-02 & 1.55E-02 & \lightbluehl{1.39E-02} & 1.42E-02 & 1.44E-02 & \bluehl{\textbf{1.31E-02}}$^{-0.18}$ \\
            \hline
            \textbf{UE} & 1.75E-02 & 2.07E-02 & 1.69E-02 & 1.84E-02 & \lightbluehl{1.63E-02} & \bluehl{\textbf{1.53E-02}}$^{-0.22}$ \\
            \hline
            \textbf{LJ} & 1.54E-02 & 1.38E-02 & 1.39E-02 & \lightbluehl{1.35E-02} & 1.40E-02 & \bluehl{\textbf{1.28E-02}}$^{-0.26}$ \\
            \hline
            \textbf{EL} & 1.45E-02 & \lightbluehl{1.17E-02} & 1.27E-02 & 1.26E-02 & 1.27E-02 & \bluehl{\textbf{1.09E-02}}$^{-0.36}$ \\
            \hline
            \hline
            \textbf{Mean} & 1.53E-02 & 1.35E-02 & 1.38E-02 & \lightbluehl{1.34E-02} & 1.41E-02 & \bluehl{\textbf{1.20E-02}}$^{-0.33}$ \\
            \hline
        \end{tabular}
    }
    \caption{DeLiVER Expected Calibration Error (ECE) performance under adverse weather and corner cases.}
    \label{tab:deliver_val_ece}
\end{table}

Table~\ref{tab:deliver_val_ece} shows the Expected Calibration Error (ECE) metric for all methods across the same scenarios.
ECE measures how well a model is calibrated by comparing it's predicted probabilities with the actual probabilities of the ground truth distribution.
ECE helps us understand how reliable a model's confidence scores are.
A higher ECE means the model is more overconfident (or unsure) about its predictions.
As shown, \ourmethod has the lowest ECE score over all methods across every weather scenario and corner case with a 0.33 average decrease compared to the baseline.
These results demonstrate that our method helps better calibrate the overconfident semantic segmentation model.
In fact all UQ methods on average has a lower ECE compared to the baseline demonstrating its usefulness in model calibration.
					
\subsection{Pre vs. Post Uncertainty Feature Learning}
\begin{table}[!ht]
    \scriptsize
    \centering
    \addtolength{\tabcolsep}{-0.3em}{
        \begin{tabular}{l|ccccc|c}
            \hline
            \multirow{3}{*}{\textbf{Scenarios}} & \multicolumn{6}{c}{\textbf{UQ Methods (Metrics: mIoU~$\uparrow$)}} \\
            \cline{2-7}
            & CMNeXt & InfMCD & 
            InfNoise & PostNet & LDU & \ourmethod \\
            & \cite{zhang2023delivering} & \cite{mi2022training} & \cite{mi2022training} & \cite{charpentier2020posterior} & \cite{franchi2022latent} & (Ours) \\
            \hline
            \textbf{Cloudy} & 68.7 & 68.78 & 68.5 & 69.13 & 69.06 & 69.26 \\
            \hline
            \textbf{Foggy} & 65.66 & 65.78 & 65.59 & 65.89 & 65.92 & 66.40 \\
            \hline
            \textbf{Night} & 62.46 & 62.7 & 62.44 & 62.82 & 63.08 & 63.71 \\
            \hline
            \textbf{Rainy} & 67.5 & 67.66 & 67.37 & 68.02 & 67.84 & 68.29 \\
            \hline
            \textbf{Sunny} & 66.57 & 66.52 & 66.39 & 66.78 & 66.78 & 67.43 \\
            \hline
            \hline
            \textbf{MB} & 62.91 & 63.11 & 62.79 & 63.40 & 63.3 & 63.78 \\
            \hline
            \textbf{OE} & 64.59 & 64.94 & 64.6 & 64.91 & 64.95 & 65.20 \\
            \hline
            \textbf{UE} & 60.00 & 59.75 & 60.07 & 60.12 & 60.3 & 60.72 \\
            \hline
            \textbf{LJ} & 65.92 & 66.14 & 66.01 & 66.17 & 66.21 & 66.44 \\
            \hline
            \textbf{EL} & 65.48 & 65.7 & 65.66 & 65.91 & 66.03 & 66.34 \\
            \hline
            \hline
            \textbf{Mean} & 66.30 & 66.59 & 66.49 & 66.60 & 66.64 & 66.98 \\
            \hline
        \end{tabular}
    }
    \caption{DeLiVER mean Intersection over Union (mIoU) performance under adverse weather and corner cases using post-fusion features.}
    \label{tab:deliver_post_fusion}
\end{table}

Table~\ref{tab:deliver_post_fusion} evaluates the performance of uncertainty feature weighting after the fusion block of Figure~\ref{fig:proposed} or see the architecture figure in the Appendix.
As expected, we found that the overall performance gain is limited/lower compared to its prefusion uncertainty fusion counterpart.
These results align with our motivation that different modalities experience different modality-specific uncertainties.
And these uncertainties can propagate to the fusion module, undermining the fusion features and degrading downstream task performance.

\subsection{Model Ablation}
\begin{table}[!ht]
    \scriptsize
    \centering
    \addtolength{\tabcolsep}{-0.4em}{
        \begin{tabular}{l|c|c}
            \hline
            \textbf{Dataset} & \underline{aiMotive} & \underline{DeLiVER} \\
            \textbf{Metric} & \textbf{all-point AP/11-point AP ~$\uparrow$} & \textbf{mIoU~$\uparrow$} \\
            \hline
            \hline
            \ourmethod & \bluehl{\textbf{66.70}/\textbf{66.00}}~(Mean $\Delta$) & \bluehl{\textbf{67.59}}~(Mean $\Delta$) \\
            \hdashline
            -- w/o Patch (4x) Proj. & 65.67/65.23 \textcolor{orange}{(-1.03/-0.77)} & 67.18 \textcolor{orange}{(-0.41)} \\
            -- w/o Chan Proj. & 64.43/64.25 \textcolor{orange}{(-1.24/-0.98)} & 66.63 \textcolor{orange}{(-0.55)}
            \\
            \hline
        \end{tabular}
    }
    \caption{Model ablation for aiMotive and DeLiVER.}
    \label{tab:ablation}
    
\end{table}
For our model ablation, we analyzed the effects of sequentially removing our proposed components and the impact on the overall performance.
As seen from Table~\ref{tab:ablation} the removal of the Patch Projection (with 4 patches) results in a 1.03/0.77 decrease in performance and the removal of Channel Projection in a 1.24/0.98 decrease in performance for aiMotive.
Similarily for the DeLiVER dataset, the removal of the Patch Projection (with 4 patches) results in a 0.41 decrease in performance and removing Channel Projection results in a 0.55 decrease in performance.
Together we see that both methods contribute significantly to the overall performance demonstrating that both channel and spatial uncertainties should be accounted for to better capture the holistic uncertainty of the features before fusion.
More ablations in the Appendix.
\subsubsection{Sensitivity Analysis}
\begin{figure}[ht]
\centering
\includegraphics[width=\linewidth]{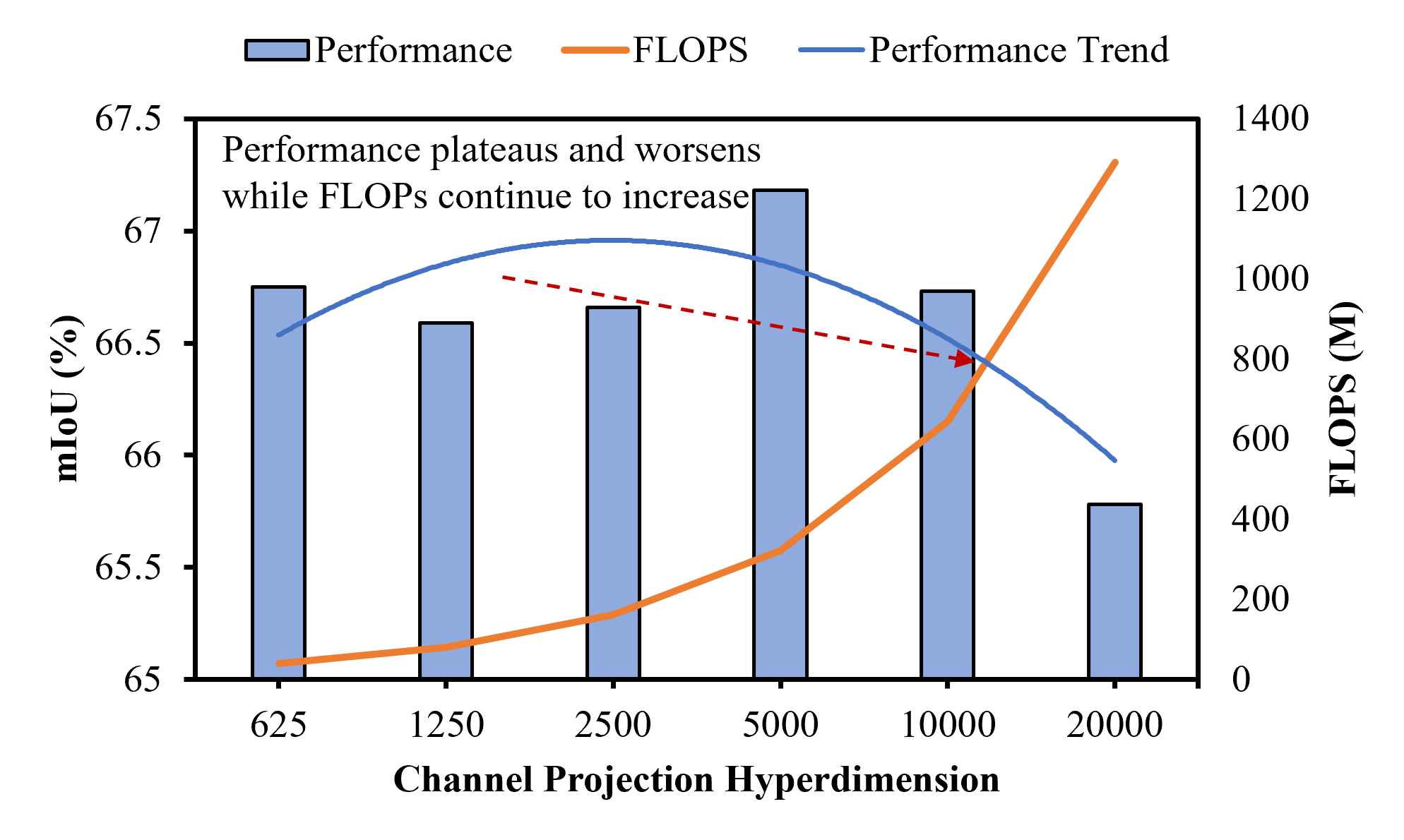}
\caption{Analysis on the effects of hyperdimension selection for channel projection on performance and computation (DeLiVER).}
\label{fig:sens_dim}
\end{figure}
Here we investigate the tunable parameters for channel and patch projection \& bundling (CPB \& PPB).
For CPB, the only parameter is the channel hyperdimension.
For PPB, we have the patch hyperdimension and the number of patches.
Figure~\ref{fig:sens_dim} shows a sweep over a range of dimensions from 625 to 20k for CPB.
This range of values is chosen based on the HDC expressivity limits (around 10k) ~\cite{yu2022understanding, thomas2021theoretical}.
We see that performance improves with dimension increase, but plateaus and decreases past 10k while computation continues to grow.
This occurs when large dimensions induce too much sparsity, making similar samples become orthogonal and capturing irrelevant information likened to over-fitting in neural networks~\cite{thomas2021theoretical}.

\begin{figure}[ht]
\centering
\includegraphics[width=\linewidth]{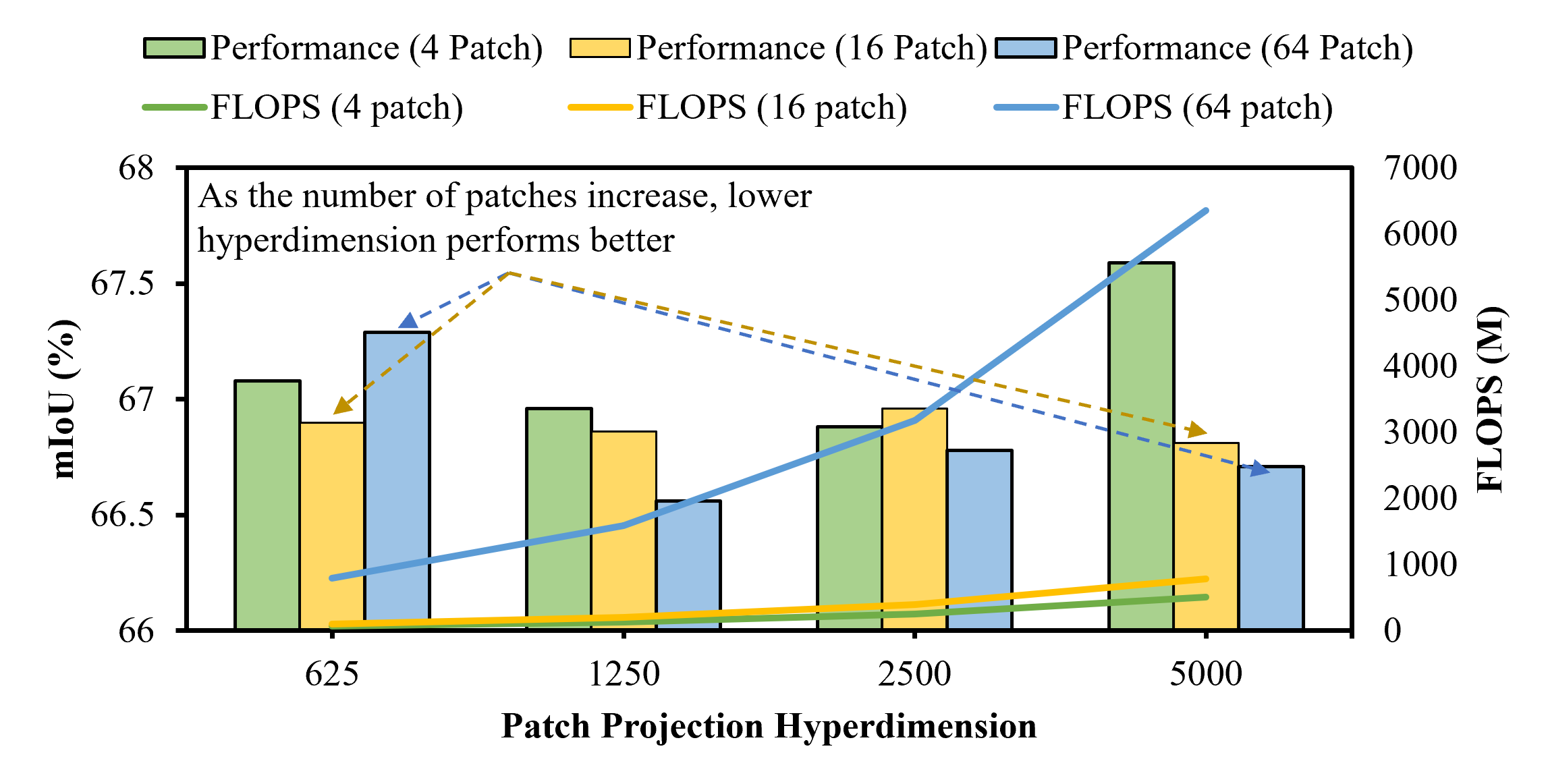}
\caption{Analysis on the effects of patches and patch hyperdimension selection on performance and computation (DeLiVER).}
\label{fig:sens_patch}
\end{figure}

Figure~\ref{fig:sens_patch} shows a sweep over 4, 16 and 64 patch configurations together with the dimension sweep from 625 to 5000.
We observe an interesting relation where the number of patches negatively correlates with the per-patch hyperdimension in terms of better performance.
This makes sense as we increase the number of patches, the information needed to be encoded becomes smaller due to the reduction in the per-patch spatial resolution as indicated in Section~\ref{sec:ppb}.
Therefore, the necessary hyperdimensions needed to encode the information should be reduced, while larger dimensions begin to over-fit. Whereas the opposite is true for less patches, larger hyperdimensions are needed.

\subsection{Computation Cost}
\begin{table}[!ht]
    \scriptsize
    \centering
    \addtolength{\tabcolsep}{-0.4em}{
        \begin{tabular}{l|ccccc}
            \hline
            \textbf{Dataset} & \multicolumn{2}{c}{\underline{aiMotive}} & & \multicolumn{2}{c}{\underline{DeLiVER}} \\
            \textbf{UQ Method} & \textbf{FLOPs} & \textbf{\#Params} & & \textbf{FLOPs} & \textbf{\#Params} \\
            \hline
            \hline
            InfMCD~\cite{mi2022training} & 990.90M & \bluehl{\textbf{4.05K}} & & 1.84G & \bluehl{\textbf{20.19K}} \\
            InfNoise~\cite{mi2022training} & 990.90M & \bluehl{\textbf{4.05K}} & & 1.84G & \bluehl{\textbf{20.19K}} \\
            LDU~\cite{franchi2022latent} & \lightbluehl{264.24M} & 44.04M & & \lightbluehl{613.42M} & 102.24M \\
            \ourmethod & \bluehl{\textbf{111.84M}} & \lightbluehl{1.15M} & & \bluehl{\textbf{338.40M}} & \lightbluehl{38.85M} \\
            \hdashline
            -- w/o Patch (4x) Proj. & 84.00M & 1.13M & & 256.00M & 38.78M \\
            -- w/o Chan Proj. & 6.96M & 3.38K & & 20.60M & 16.83K \\
            \hline
        \end{tabular}
    }
    \caption{Comparing inference (FLOPs) and training costs (\# of Trainable Parameters).}
    \label{tab:computation}
\end{table}

Finally, the main benefit of \ourmethod is its efficiency for training and inference.
We measure the computation costs in terms of Floating point operations (FLOPs) and the number of trainable model parameters in Table~\ref{tab:computation}.
We approximate the FLOPs by computing the major contributing operations.
Refer to the Appendix for our FLOPs approximations for each operation type.
For InfMCD and InfNoise these are the entropy, mutual information, and predictive entropy computations.
For LDU, it is the cosine similarity between prototypes and matrix multiplication for uncertainty, which was found to be negligible.
For \ourmethod, it is the input projection and cosine similarity computations.
We see that \ourmethod is up to 2.36$\times$ and 1.81$\times$ more FLOP-efficient compared to LDU for aiMotive and DeLiVER respectively.
Compared to traditional Bayesian approximation methods, it can be up to 8.86$\times$ and 5.44$\times$ more FLOP-efficient.
As for the training costs, \ourmethod has 38.30$\times$ and 2.63$\times$ less trainable parameters compared to LDU for aiMotive and DeLiVER respectively, making it more memory efficient.
This ultimately shows the practicality of \ourmethod as a UQ method for real-world autonomous systems.

%% file: chapters/5_conclusion.tex
\section{Discussions and Conclusion}\label{sec:conclusion}
\noindent\textbf{Limitations and Future Work}:
We leveraged cloud computing resources using a single NVIDIA A100 GPU for training.
Our method uses the traditional bundling technique, which requires labels for supervised learning of prototypes.
The need for labels is the primary limitation of our approach.
Although there have been implementations of semi-supervised and unsupervised versions \cite{imani2019hdcluster, imani2019semihd}, there are no theoretical guarantees and/or justifications for their performance unlike the traditional bundling method \cite{thomas2021theoretical}.
We leave the exploration of these methods for uncertainty quantification to future work. 
Additionally, the FLOPs of the vector symbolic architectures may be further reduced due to the massively parallel operations of hyperdimensional computation. 
These optimizations were demonstrated in prior works  \cite{imani2020dual, gupta2020thrifty}.

\noindent\textbf{Conclusion}:
In this work, we propose an efficient method to quantify the epistemic uncertainty of machine learning models at the feature level.
\ourmethod projects the latent features as hypervectors to form hyperdimensional prototypes.
Similarity between hypervectors of new samples and learned prototypes is used as a proxy to estimate uncertainty.
We demonstrate the practicality of \ourmethod through autonomous driving tasks, including object detection and semantic segmentation.
Our results show that our method can enhance the performance and robustness of multimodal fusion models under uncertainties induced by various weather conditions and modality-specific corner cases.
We achieve this by quantifying the feature uncertainty of each modality and applying an uncertainty-aware weighting layer prior to feature fusion and fine-tuning the post-fusion layers.
Finally, we demonstrated that \ourmethod requires significantly less FLOPs and training parameters when compared to existing works, indicating its real-world practicality.

\noindent\textbf{Acknowledgements:}
This work acknowledges the support of the Automotive Research Center (ARC), Cooperative Agreement W56HZV-24-2-0001 U.S. Army DEVCOM Ground Vehicle Systems Center (GVSC).
\clearpage

%% file: include/appendix.tex
\clearpage
\setcounter{page}{1}
\maketitlesupplementary

\section{Similarity to Uncertainty/Expressivity to Uncertainty Quantifiability}
The projection function $\phi$ is an encoding from $\mathbb{R}^n \rightarrow \mathbb{R}^d$ that can be mathematically expressed as:
\begin{equation}\label{eqn:proj}
\phi(z) = \Phi\cdot z
\end{equation}
where $\Phi\in\mathbb{R}^{d\times n}$ is the projection matrix and the output hypervector space $\mathcal{\textbf{H}}$ is an \textbf{inner-product space}.

According to \cite{yu2022understanding}, the initialization of $\Phi$ can limit the learnability of a VSA system.
They proved this by defining VSA systems by their \textbf{expressivity} as follows:
\begin{definition}\label{def:expressibility}
A VSA system can express a similarity matrix $\mathcal{M} \in \mathbb{R}^{N\times N}$ if for any $\epsilon > 0$, there exists a $d \in \mathbb{N}$ and $d$-dimensional hypervectors $\mathcal{H}_{1}, \mathcal{H}_{2}, \cdots, \mathcal{H}_{N}$ such that $|\mathcal{M}_{i,j} - \delta(\mathcal{H}_{i}, \mathcal{H}_{j})|\leq\epsilon$
\end{definition}
Since $\mathcal{M}$ represents the knowledge of all sample relations in the $\mathcal{\textbf{H}}$ space, the expressivity depends on whether $\phi$ can capture the similarities of $\mathcal{M}$ accurately.
Given the knowledge of $\mathcal{M}$,  \textbf{we derive an expression for the \textit{uncertainty quantifiability} of a VSA system as follows:}
\begin{corollary}\label{def:uncertainty}
A VSA can express an uncertainty similarity matrix $\mathcal{U} \in \mathbb{R}^{L\times N}$ if for any $\eta > 0$, there exists a $d \in \mathbb{N}$ and $d$-dimensional hypervectors $\mathcal{H}_{1}, \mathcal{H}_{2}, \cdots, \mathcal{H}_{N}$ such that $|\mathcal{U}_{i,j} - \delta(\mathcal{H}_{i}, \mathcal{H}_{j})|\leq\eta$. 
\end{corollary}
In this Corollary, the rows of the uncertainty similarity matrix $\mathcal{U}$ are the prototypes $\mathcal{H}^{l}_{m}$ (Equation \ref{eqn:hdp}) and the columns are the similarities of each hypervectors $\mathcal{H}_{1}, \mathcal{H}_{2}, \cdots, \mathcal{H}_{n}$ to each $\mathcal{H}^{l}_{m}$ (Equation \ref{eqn:uncert1}).

Expressivity thus directly impacts a VSA's ability to accurately quantify uncertainty.
\footnote{Obtaining $\mathcal{M}$ involves solving an intractable linear programming problem
of size exponential in $N$, making it unrealistic both from a computation and memory perspective for arbitrarily large datasets.}
\cite{yu2022understanding} showed that classical initializations using the kernel trick \cite{rahimi2007random} has limited expressiveness.
Instead, Random Fourier Features (RFF) can, in expectation, exactly achieve $M$ or some approximation of $M$.

However, the above assumes that the \textbf{orthogonality} between hypervectors is maintained by $\phi$.
When assessing whether some $\mathcal{H}_{z} \in \mathcal{H}^{l}_{m}$ and the encodings are perfectly orthogonal, we get the expression $|\delta(\mathcal{H}_{z}, \mathcal{H}^{l}_{m})| = I\mathbbm{1}(\mathcal{H}_{z} \in \mathcal{H}^{l}_{m})$ where $\mathbbm{1}$ is an indicator that evaluates to one if true and zero otherwise, $I=\min_{z\in \mathcal{Z}}||\phi(z)||^2$ when $\delta$ is the dot-product or $I=1$ when $\delta$ is the cosine similarity.

According to \cite{thomas2021theoretical}, when orthogonality is not maintained, it can cause interference in the hypervector encoding.
They characterizes this as the \textbf{incoherence} which limits the expressivity of $\phi$ as follows:
\begin{definition}\label{def:incoherence}
    For $\mu\geq0, \phi $ is $\mu$-incoherent if for all distinct $z,z' \in \mathcal{Z}$ we have
    \begin{equation}\label{eqn:incoherence}
        |\delta(\phi(z), \phi(z'))| \leq \mu I
    \end{equation}
\end{definition}

When the encoding $\phi(z)$ and $\phi(z')$ are \textbf{not perfectly orthogonal} ($\mu=0$), the interference causes a $\Delta$ ``cross-talk'' such that $\delta(\mathcal{H}_{z},\mathcal{H}^{l}_{m}) = I\mathbbm{1}(\mathcal{H}_{z}\in \mathcal{H}^{l}_{m}) + \Delta$.
\textbf{By combining Corollary \ref{def:uncertainty} \& Definition \ref{def:incoherence} we propose a new definition for the \textit{uncertainty quantifiability of a $\mu$-incoherent $\phi$} as follows:}
\begin{definition}\label{def:uncer_incoh}
    A $\mu$-incorherent VSA system can express an uncertainty matrix $\mathcal{U} \in \mathbb{R}^{L\times N}$ if for any $\eta, \mu > 0$, there exists a $d \in \mathbb{N}$ and $d$-dimensional hypervectors $\mathcal{H}_{1}, \mathcal{H}_{2}, \cdots, \mathcal{H}_{N}$ s.t. $|\mathcal{U}_{i,j}|\leq\mu I+\eta$
\end{definition}

Thus, to reliably quantify uncertainty, we must ensure the contribution of $\mu$ is small.
Given that the loss of orthogonality contributes to incoherence, the instinct is to improve $\phi$ by enforcing orthogonality onto the initialization of the projection matrix $\Phi$.

\section{Proof of Equation} \label{apndx:uncer_incoh}

Given the following inequalities:
\begin{align}
\setcounter{equation}{0}
|\mathcal{U}_{i,j} - \delta(\mathcal{H}_{i}, \mathcal{H}_{j})|\leq\eta \label{eqn:uncer_incoh_1} \\
|\delta(\mathcal{H}_{i}, \mathcal{H}_{j})|\leq\mu L \label{eqn:uncer_incoh_2}
\end{align}
\textbf{Step 1.} Rewrite (\ref{eqn:uncer_incoh_1}) \& (\ref{eqn:uncer_incoh_2}) as follows:
\begin{align}
-\eta \leq \mathcal{U}_{i,j} - \delta(\mathcal{H}_{i}, \mathcal{H}_{j})\leq \eta \label{eqn:uncer_incoh_1_re} \\
-\mu L \leq \delta(\mathcal{H}_{i}, \mathcal{H}_{j})\leq\mu L \label{eqn:uncer_incoh_2_re}
\end{align}
\textbf{Step 2.} Add $\delta(\mathcal{H}_{i}, \mathcal{H}_{j})$ to both sides of (\ref{eqn:uncer_incoh_1}):
\begin{align}
\delta(\mathcal{H}_{i}, \mathcal{H}_{j})-\eta \leq \mathcal{U}_{i,j}\leq \delta(\mathcal{H}_{i}, \mathcal{H}_{j})+\eta \label{eqn:uncer_incoh_1_re2}
\end{align}
\textbf{Step 3.} Define upper and lower bounds for (\ref{eqn:uncer_incoh_2_re}) and (\ref{eqn:uncer_incoh_1_re2}):

\textbf{Upper Bounds}:
\begin{align}
\delta(\mathcal{H}_{i}, \mathcal{H}_{j})\leq\mu L\\
\mathcal{U}_{i,j}\leq \delta(\mathcal{H}_{i}, \mathcal{H}_{j})+\eta
\end{align}

\textbf{Lower Bounds}:
\begin{align}
-\mu L \leq \delta(\mathcal{H}_{i}, \mathcal{H}_{j})\\
\delta(\mathcal{H}_{i}, \mathcal{H}_{j})-\eta \leq \mathcal{U}_{i,j}
\end{align}

\textbf{Step 4.} Substitute terms and rewrite bounds:

\textbf{Upper Bounds}:
\begin{align}
\mathcal{U}_{i,j}\leq \mu L+\eta
\end{align}

\textbf{Lower Bounds}:
\begin{align}
-\mu L-\eta \leq \mathcal{U}_{i,j}
\end{align}

\textbf{Step 5.} Rewrite inequality, end of proof:
\begin{align}
-(\mu L+\eta) \leq \mathcal{U}_{i,j}\leq \mu L+\eta = |\mathcal{U}_{i,j}| \leq \mu L+\eta
\end{align}

\section{FLOPs Computation Approximation}\label{apndx:flops}
We approximate the number of Floating-Point Operations for each method by deriving the major contributing operations as follows:

\textbf{Matrix Multiplication FLOPs Approximation:}
Projection encoding uses the dot product operation which involves matrix multiplication and addition.
We assume an input size of $In=1\times C$, a projection matrix size of $Proj=C\times D$, and an expected output size of $Out=B\times D$. 
For each element $Out_{i,j}$ there are $C$ multiplications.
For each element $Out_{i,j}$ there are $C-1$ additions (because we need to add $C$ products together, which requires $C-1$ additions).
Where $B$ is the batch size (we are assuming B=1 for the sake of simplicity), $C$ is the channel size, and $D$ is the output dimension/hyperdimension.
Thus, for the entire output $Out$ with $B\times D$ elements:
\begin{align}
\text{Total Multiplications} &= B\cdot D\cdot C \\ 
\text{Total Additions} &= B\cdot D\cdot(C-1) \\ 
\text{Total FLOPs} &= B\cdot D\cdot C \\ &+ B\cdot D\cdot(C-1) \notag\\ 
            &= 2\cdot B\cdot D\cdot C - B\cdot D \\ 
            &= B\cdot D(2\cdot C-1) \\ &\approx 2\cdot B\cdot D\cdot C 
\end{align}
\textbf{Einstein Summation FLOPs Approximation}: is a general case of matrix multiplication where we used it for our Channel-Projection method.
Specifically, Einsum(“BC,CD->BCD”, A, B), which reduces the computation to an outer product, obtains $\text{Out}=B\times C\times D$ with 0 addition FLOPs:
\begin{align}
\text{Total Multiplications} &= B\cdot D\cdot C \\ 
\text{Total Additions} &= 0 \\ 
\text{Total FLOPs} &= B\cdot D\cdot C
\end{align}

\textbf{Cosine Similarity FLOPs Approximation:}
$cosine\_{similarity}(u,v) = \frac{u\cdot v}{||u||||v||}$ involves dot product along with division and two Euclidean norm operations.
We assume an input $u$ of size $B\times D$ and the comparison $v$ is of size $1\times D$.
Since $B=1$, the dot product is between two vectors, thus involving $D$ multiplications and $D-1$ additions.
For Euclidean Norm $||u|| = \sqrt{\sum^{D}_{d=1}u^{2}_{d}}$, squaring involves $D$ multiplications, summing involves $D-1$ additions, and we assume square root is 1 operation, thus the total FLOPs for cosine similarity is:
\begin{align}
    \text{Dot Product FLOPs} &= D + (D-1) \\
    &= 2\cdot D-1 \\
    \text{Euclidean Norm FLOPs} &= D + (D-1) + 1 \\
    &= 2\cdot D \\
    \text{Division + Norm Mult. FLOPs} &= 2 \\
    \text{Total FLOPs} &= (2\cdot D + 2\cdot D \\
    & + 2\cdot D + 2) \notag \\
    &= 6\cdot D + 1 \approx 6\cdot D
\end{align}

\textbf{Entropy FLOPs Approximation:}
The deterministic entropy equation $H(P) = -\sum_{i}P(i)logP(i)$ involves addition from summation, multiplication, and logarithm.
We assume that $P$ has dimensions $B\cdot C \cdot H \cdot W$, and thus the number of addition, multiplication, and logarithmic operations depends on the number of elements which is $B\cdot C \cdot H \cdot W$.
Therefore, the total number of FLOPs for Entropy is:
\begin{align}
    \text{Total elements} &= B\cdot C \cdot H \cdot W \\
    \text{Mult. + Log FLOPs} &= 1 + 1 = 2\\
    \text{Sum FLOPs} &= B\cdot C \cdot H \cdot W \\
        \text{Total FLOPs} &= \text{2}\cdot\text{Tot. elems.} \\
        &+ \text{Sum FLOPs} \notag\\
        &= 3 \cdot B\cdot C \cdot H \cdot W
\end{align}
We note that predictive entropy and mutual information are more complex operations that induce more FLOPs (refer to \cite{tian2020uno}), for simplicity, we assume the same FLOPs as the deterministic entropy. 

From here, we demonstrate the aiMotive FLOPs computations in Table~\ref{tab:computation} as an example with $B=1$.
The BEVFusion model from aiMotive generates RGB features (BEVDepth) and Lidar+Radar (VoxelNet) features where the total feature dimensions together are $C\times H\times W = 336\times 64 \times 512$ (RGB = 80 channels and Lidar+Radar = 256).

\textbf{InfMCD}: involves 10 forward passes to generate feature outputs (realistically all previous computation layer FLOPs in the model should be accounted for), followed by predictive entropy, mutual information, and deterministic entropy computations (we assume 3$\times$ deterministic entropy FLOPs) \\
\begin{align}
    \text{Total elements} &= 336\times 64\times 512 = 11,010,048 \\
    \text{Total FLOPs} &= 10~\text{outputs} \times 3\times~\text{Entropy FLOPs} \\
    &= 30 \times 3\times 11,010,048 \\
    &= 990,904,320 = 990.90~\text{MFLOPs}
\end{align}
\textbf{InfNoise}: involves 10 forward passes to generate Gaussian noise added features (realistically FLOPs for generating Gaussian noise and adding noise to the input of a specified layer along with  FLOPs for obtaining the new layer outputs should be accounted for), followed by predictive entropy, mutual information, and deterministic entropy computations (we assume 3$\times$ deterministic entropy FLOPs).
This results in the same approximated FLOPs as InfMCD.

\textbf{LDU}: involves comparing the cosine\_similarity of the features with the learned prototypes along with a Conv2d for uncertainty estimation (ignored due to negligible cost).
Since LDU allows arbitrary prototypes, we set it to $L=4$, the same number as our method based on the number of aiMotive scenarios.
Additionally, LDU's prototypes are defined with the shape $1\times L\times C\times H\times W$.
\begin{align}
    \text{Vector $u$ dimension} &= 336\times 64\times 512 \\ 
    &= 11,010,048 \\
    \text{Total FLOPs} &= L \times \text{Cos\_Sim FLOPs}\\
    &= 4\times 6\times u \\
    &= 24 \times 11,010,048 \\
    &= 264,241,152 \\
    &= 264.24~\text{MFLOPs}
\end{align}

\textbf{\ourmethod}: involves the feature projection and cosine similarity computation, where we set $d=10k$ as real-valued hyperdimensional prototypes and $L=4$.
Particularly, for the Einsum Matrix multiplication, the 
\begin{align}
    \text{Vector $u$ dim} = 1 \times d = 10k \\
    \text{Tot. FLOPs w/o Chan. Proj.} = \text{MatMul FLOPs} \\
    +~L \times \text{Cos\_Sim FLOPs} \notag \\
    = 2\times d \times 336 + 4\times 6\times u \\
    = 2\times 10k \times 336 + 24 \times 10k \\
    = 6.96~\text{MFLOPs} \\
    \text{Vector $u$ Chan. Proj. dim} = 1\times c\times d = 3.36M \\
    \text{Tot. FLOPs w/ Chan. Proj.} = \text{Einsum FLOPs}~+ \\
    L \times \text{Cos\_Sim FLOPs} \notag 
 \\
    = d \times c + 4\times 6\times u \\
    = 10k\times 336 + 24 \times 3.36M \\
    = 84.00~\text{MFLOPs}
\end{align}
\ourmethod FLOPs breakdown by component for aiMotive: 4x (patches) spatial projection=4*6.72M=26.88M, 4x spatial similarity=4*240K=960K, channel projection=3.36M, channel similarity=80.64M (\textbf{72\% of all costs}).
Uncertainty weights=(kern\_h*kern\_w*in\_c+1)*(out\_h*out\_w*out\_c) =(4*1*336+1)*(1*1*336)=452K. (Computation computed for Conv layer described in Architectures figure, approx.~same for all methods, thus ignored).

\clearpage
\section{Architectures}
\begin{figure}[ht]
\centering
\onecolumn\includegraphics{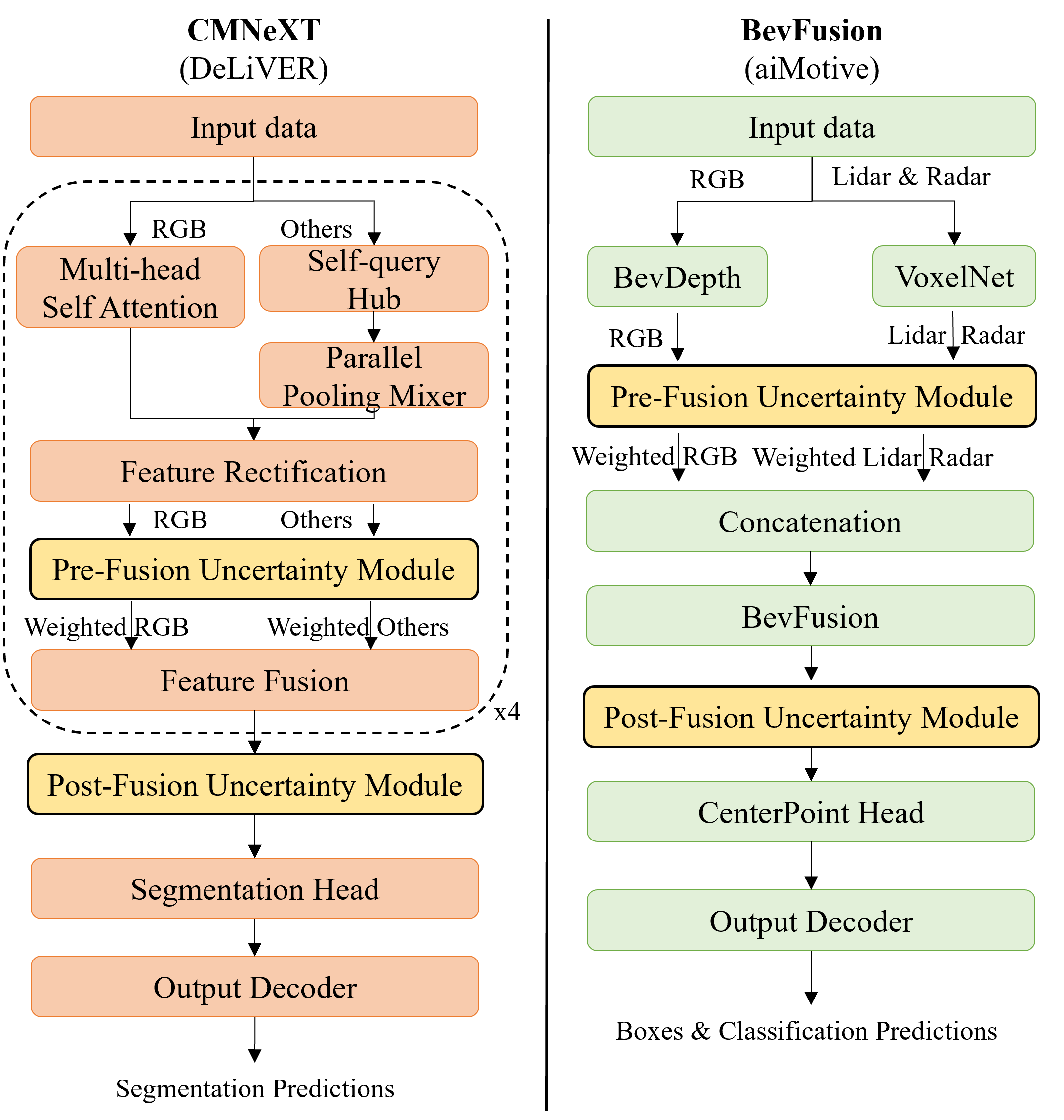}
\caption{Architectures diagram showing where we insert the uncertainty module for pre and post fusion methods. 
The uncertainty weighting is a single Conv layer, Input: (B,M,P,C)/(B,M,P,1), kernel=(P,1), stride=(P,1), Output: (B,M,1,C)/(B,M,1,1) for channel/patch weights respectively.
B=Batch, M=Modality, P=Prototype, C=Channel.
\underline{Channel}/\textbf{Patch} weights are multiplied uniformly across all \underline{spatial}/\textbf{channel} dimensions per \underline{channel}/\textbf{patch}.
The weighting module performs dimension matching automatically.}
\label{fig:architectures}
\end{figure}
\twocolumn

\clearpage
\section{Additional Experiments}
\begin{table}[ht]
    \onecolumn
    \centering
    \addtolength{\tabcolsep}{-0.4em}{
        \begin{tabular}{l|ccccc|ccccc|c}
            \hline
            \textbf{UQ Method} & \textbf{Cloudy} & \textbf{Foggy} & \textbf{Night} & \textbf{Rainy} & \textbf{Sunny} & \textbf{MB} & \textbf{OE} & \textbf{UE} & \textbf{LJ} & \textbf{EL} & \textbf{Mean} \\
            \hline
            CMNeXt~\cite{zhang2023delivering} & 53.05 & 54.06 & 50.63 & 54.26 & 51.88 & 49.25 & 49.40 & 47.08 & 52.28 & 53.62 & 53.00 \\
            InfMCD~\cite{mi2022training} & \lightbluehl{53.67} & \lightbluehl{54.83} & \lightbluehl{51.42} & \lightbluehl{54.72} & \bluehl{\textbf{52.37}} & \lightbluehl{49.67} & 51.25 & \lightbluehl{51.18} & \lightbluehl{53.02} & \bluehl{\textbf{54.02}} & \lightbluehl{53.41} \\
            InfNoise~\cite{mi2022training} & 53.25 & 54.37 & 50.85 & 54.52 & 51.95 & 49.39 & 51.08 & 50.86 & 52.83 & 53.37 & 53.19 \\
            PostNet~\cite{charpentier2020posterior} & 53.38 & 54.10 & 51.13 & 54.39 & 51.89 & 49.48 & 51.15 & 50.99 & 52.68 & 53.20 & 53.15 \\
            LDU~\cite{franchi2022latent} &  53.66 & 54.40 & 51.41 & 54.46 & 51.94 & 49.38 & 51.32 & 51.04 & 52.92 & 53.31 & 53.37 \\
            \ourmethod & \bluehl{\textbf{53.77}} & \bluehl{\textbf{54.91}} & \bluehl{\textbf{51.52}} & \bluehl{\textbf{54.75}} & \bluehl{\textbf{52.37}} & \bluehl{\textbf{49.69}} & \bluehl{\textbf{51.38}} & \bluehl{\textbf{51.42}} & \bluehl{\textbf{53.17}} & \lightbluehl{53.78} & \bluehl{\textbf{53.69}} \\
            \hline
        \end{tabular}
    }
    \caption{DeLiVER test set with adverse weather and corner cases: Lidar-Jitter (\textbf{LJ}), Event Low-resolution (\textbf{EL}). (Metrics: mIoU)}
    \label{tab:deliver_test}
\end{table}
\begin{table}[ht]
    \centering
    \addtolength{\tabcolsep}{-0.4em}{
        \begin{tabular}{l|cccc|l}
            \hline
            \textbf{UQ Method} & \textbf{Highway} & \textbf{Urban} & \textbf{Night} & \textbf{Rain} & \textbf{Mean ($\Delta$)} \\
            \hline
            \hline
            \ourmethod & \bluehl{\textbf{72.23/69.58}} & \bluehl{\textbf{64.77}/\textbf{64.48}} & \bluehl{\textbf{76.69}/\textbf{75.15}} & \bluehl{\textbf{44.78/45.48}} & \bluehl{\textbf{66.70}/\textbf{66.00}} \\
            \hdashline
            -- w/o Patch (4x) Proj. & 70.74/68.55 & 64.35/64.15 & 75.53/73.83 & 40.09/41.69 & 65.67/65.23 \textcolor{orange}{(-1.03/-0.77)} \\
            -- w/o Chan Proj. & 70.62/68.94 & 62.03/59.90 & 75.11/74.04 & 34.01/36.21 & 64.43/64.25 \textcolor{orange}{(-1.24/-0.98)} \\
        \end{tabular}
    }
    \caption{aiMotive ablation under diverse scenes. (Metrics: all-point AP/11-point interpolation AP)}
    \label{tab:aimotive_scenes_ablation}
\end{table}
\begin{table}[ht]
    \centering
    \addtolength{\tabcolsep}{-0.4em}{
        \begin{tabular}{l|cccc|l}
            \hline
            \textbf{UQ Method} & \textbf{MB} & \textbf{OE} & \textbf{UE} & \textbf{LF} & \textbf{Mean ($\Delta$)} \\
            \hline
            \hline
            \ourmethod & \bluehl{\textbf{64.39}/\textbf{63.99}} & \bluehl{\textbf{66.16}/\textbf{65.39}} & \bluehl{\textbf{64.18}/\textbf{63.91}} & \bluehl{\textbf{65.89}/\textbf{65.22}} & \bluehl{\textbf{65.16/64.62}} \\
            \hdashline
            -- w/o Patch (4x) Proj. & 63.58/63.35 & 65.26/64.62 & 63.38/63.40 & 64.85/64.42 & 64.27/63.95 \textcolor{orange}{(-0.89/-0.67)} \\
            -- w/o Chan Proj. & 62.65/62.43 & 64.76/63.16 & 62.29/62.22 & 63.60/63.02 & 63.47/62.83 \textcolor{orange}{(-0.80/-0.88)}\\
        \end{tabular}
    }
    \caption{aiMotive ablation under corner cases. (Metrics: all-point AP/11-point interpolation AP)}
    \label{tab:aimotive_corner_ablation}
\end{table}
\begin{table}[ht]
    \centering
    \addtolength{\tabcolsep}{-0.4em}{
        \begin{tabular}{l|ccccc|ccccc|l}
            \hline
            \textbf{UQ Method} & \textbf{Cloudy} & \textbf{Foggy} & \textbf{Night} & \textbf{Rainy} & \textbf{Sunny} & \textbf{MB} & \textbf{OE} & \textbf{UE} & \textbf{LJ} & \textbf{EL} & \textbf{Mean ($\Delta$)} \\
            \hline
            \hline
            \ourmethod & \bluehl{\textbf{53.77}} & \bluehl{\textbf{54.91}} & \bluehl{\textbf{51.52}} & \bluehl{\textbf{54.75}} & \bluehl{\textbf{52.37}} & \bluehl{\textbf{49.69}} & \bluehl{\textbf{51.38}} & \bluehl{\textbf{51.42}} & \bluehl{\textbf{53.17}} & \bluehl{\textbf{53.78}} & \bluehl{\textbf{53.69}} \\
            \hdashline
            -- w/o Patch (4x) Proj. & 53.22 & 54.60 & 51.16 & 54.69 & 52.07 & 49.69 & 51.25 & 51.16 & 53.13 & 53.69 & 53.37 \textcolor{orange}{(-0.32)} \\
            -- w/o Chan Proj. & 53.27 & 54.44 & 50.87 & 54.31 & 51.83 & 49.47 & 51.02 & 50.54 & 52.65 & 53.47 & 53.15 \textcolor{orange}{(-0.54)} \\
        \end{tabular}
    }
    \caption{DeLiVER test set ablation with adverse weather and corner cases. (Metrics: mIoU)}
    \label{tab:deliver_ablation_appdx}
\end{table}

\subsection{\#Prototypes \& Held-Out Scenarios}
\begin{table}[!ht]
    \onecolumn
    \centering
    \addtolength{\tabcolsep}{-0.4em}{
        \begin{tabular}{c|c|ccccc|c}
            \hline
            \textbf{Experiment} & \textbf{UQ Method} & \textbf{Cloudy} & \textbf{Foggy} & \textbf{Night} & \textbf{Rainy} & \textbf{Sunny} & \textbf{Mean} \\
            \hline
            Baseline & CMNeXt & 68.70 & 65.66 & 62.46 & 67.50 & 66.57 & 66.30 \\
            \hline
            \# Prototypes: 3 & \ourmethod & 69.50 & 66.00 & 63.35 & 68.28 & 67.17 & 66.94 \\
            \hline
             \# Prototypes: 10 & \ourmethod & \bluehl{\textbf{70.04}} & \bluehl{\textbf{66.48}} & \bluehl{\textbf{64.07}} & \bluehl{\textbf{68.85}} & \bluehl{\textbf{67.67}} & \bluehl{\textbf{67.53}} \\
            \hline
            \hline
            \multirow{5}{*}{Held-Out Scenarios} & InfMCD & 68.61 & 65.93 & 62.37 & 67.12 & \lightbluehl{66.90} & 66.29 \\
            \cline{2-8}
            & InfNoise & \bluehl{\textbf{69.59}} & 65.66 & \lightbluehl{62.93} & 68.13 & \bluehl{\textbf{66.95}} & \lightbluehl{66.75} \\
            \cline{2-8}
            & PostNet & 69.10 & \lightbluehl{65.91} & 62.37 & \lightbluehl{68.10} & 66.52 & 66.50 \\
            \cline{2-8}
            & LDU & \lightbluehl{69.46} & 65.36 & 62.49 & 67.68 & 66.89 & 66.48 \\
            \cline{2-8}
            & \ourmethod & 69.27 & \bluehl{\textbf{66.08}} & \bluehl{\textbf{63.75}} & \bluehl{\textbf{68.15}} & 66.81 & \bluehl{\textbf{66.91}} \\
            \hline
        \end{tabular}
    }
    \caption{Varied prototypes and held-out experiments (DeLiVER).}
    \label{tab:rebuttal_exp}
\end{table}
\twocolumn
Table~\ref{tab:rebuttal_exp} aims to evaluate the impact on performance of \ourmethod under two non-ideal conditions.
1) What would happen if we have poorly-defined scenarios, facilitating prototype definition under non-ideal conditions, i.e. fewer prototypes (e.g., 2 prototypes for 4/5 scenarios) and more prototypes for extremely specific data attributes.
2) What would happen if \ourmethod is trained without a scenario in the dataset, prototypes are computed only for the available scenarios (training data), and then the model is evaluated with samples from the held-out scenario?
To test these scenarios for fewer prototypes (protos) we merged: $cloud\_fog$, $night\_rain$, $sun$.
For more protos, we split by maps (1,2,3,6): $cloud_{1\_2}$, $cloud_{3\_6}$, $fog_{1\_2}$, $fog_{3\_6}$, $\cdots$.
Table~\ref{tab:rebuttal_exp} shows that fewer protos by \textbf{careless merging} achieve suboptimal performance, while fine-grained protos improve performance.
Merging different underlying uncertainty attributes (cloud/fog) introduces ambiguity (what is a cat/dog hybrid?).
Protos should be formed by samples that share all underlying uncertainty attributes.
\textbf{Praticality:} Well-curated datasets, i.e., nuScenes, Waymo, KITTI, etc. all capture meta-data facilitating prototype definition.
With no/limited meta-data, one can apply unsupervised methods (i.e. clustering) to identify prototypes or generate new prototypes using diffusion models.
\textbf{Held-out Scenarios:} Average performance drops as expected, but compared to others, \ourmethod maintains competitive performance.
Particularly in more uncertain (fog/night/rain) held-out scenarios.

\subsection{Uncertainty Visualization}
\begin{figure}[!ht]
  \centering
  \onecolumn\includegraphics[width=1\linewidth]{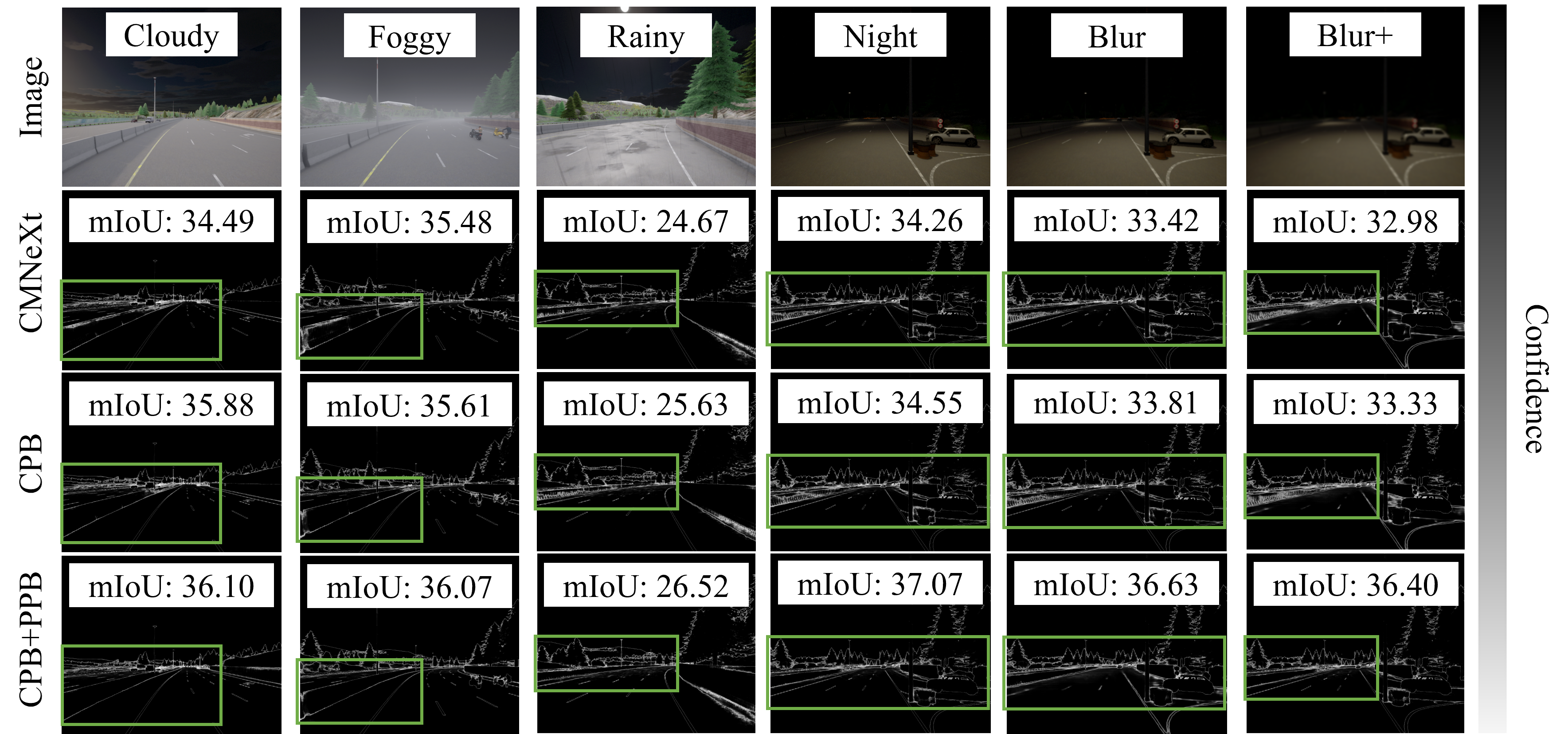}
   \caption{Channel/Patch Projection \& Bundling (CPB/PPB) effects visualization on prediction uncertainty map (DeLiVER).}
   \label{fig:uncert_vis}
\end{figure}
\twocolumn
Figure~\ref{fig:uncert_vis} ablates the channel and spatial uncertainty weighting performance through the prediction uncertainty map across scenarios and noise effects (Gaussian Blur 7x7 and 25x25).
Generally, channel sharpens and spatial refines the uncertainty maps in-detail.
Additionally, we see that CPB and PPB together improve the overall robustness to Gaussian blur noise when compared to the baseline model.

\subsection{Fine-Tuning Details}
\begin{table}[!ht]
    \onecolumn
    \centering
    \addtolength{\tabcolsep}{-0.4em}{
        \begin{tabular}{l|c|c}
        \hline
        \textbf{Hyper-parameter} & aiMotive & DeLiVER \\
        \hline
        Architecture & BEVFusion & CMNeXt \\
        \hline
        backbone & ResNet & CMNeXt-B2\\
        \hline
        learning rate & 1e-3/64& 6e-5 \\
        \hline
        batch size & 1 & 1 \\
        \hline
        epochs & 200 & 200\\
        \hline
        EarlyStopping (epochs) & 10 & 10 \\
        \hline
        weight decay & 1e-7 & 0.01 \\
        \hline
        Scheduler & MultiStepLR & warmuppolylr \\
        \hline
        Prototypes & 4 & 5 \\
        \hline
        Forwards (InfMCD \& InfNoise) & 10 & 10 \\
        \hline
        Projection Type & Ortho. Rand. Fourier Proj. & Ortho. Rand. Fourier Proj. \\
        \hline
        Channel Hyperdimension & 10000 & 5000 \\
        \hline
        Patch Hyperdimension & 10000 & 5000 \\
        \hline
        Number of Patches & 4 & 4 \\
        \hline
        \end{tabular}
    }
    \caption{Hyper-parameter configuration used for aiMotive and DeLiVER Fine-tuning. The projection function uses Random Fourier Features (rfflearn python) similar to \cite{yu2022understanding} which has no learned parameters. Prototype formation uses the train set (100\%) but can be a subset (importance selection), with \textbf{only one pass through train data} instead of multi-epochs.}
    \label{tab:aimotive_hyperparams}
\end{table}

\subsection{Artifical Noise Injection Paramters}
\begin{table}[!ht]
    \onecolumn
    \centering
    \addtolength{\tabcolsep}{-0.4em}{
        \begin{tabular}{l|c}
        \hline
        \textbf{Noise Injection} & Configuration Setting \\
        \hline
        Over Exposure & rescale\_intensity(data, in\_range=(0, int(np.max(data)/2)), out\_range=(0,255)).astype(uint8) \\
        \hline
        Under Exposure & rescale\_intensity(data, in\_range=(int(np.max(data)/2), int(np.max(data))), out\_range=(0,255)).astype(uint8) \\
        \hline
        Motion Blur & GaussianBlur(kernel\_size=(25,25), sigma=16) \\
        \hline
        Foggification & BetaRandomization(beta=1e-5) \\
        \hline
        \end{tabular}
    }
    \caption{Configuration for aiMotive artificial noise injections. GaussianBlur from torchvision.transforms, rescale\_intensity from skimage.exposure, for foggification refer to \cite{bijelic2020seeing}. These injections are only performed on the test set during evaluation and never seen during training/validation.}
    \label{tab:aimotive_noise_config}
\end{table}
\twocolumn